\documentclass[10pt,twocolumn,twoside]{IEEEtran} 

\usepackage{amsmath,amsthm,amssymb}
\usepackage{graphicx,epsfig}
\usepackage[linesnumbered,boxed]{algorithm2e}
\usepackage[caption=false,font=footnotesize]{subfig}

\def\i{i}
\def\j{j}
\def\Z{\mathbb{Z}}

\newtheorem{theorem}{Theorem}[section]
\newtheorem{proposition}[theorem]{Proposition}
\newtheorem{corollary}[theorem]{Corollary}

\begin{document}

\title{Fast and Provably Accurate Bilateral Filtering}
\author{Kunal~N.~Chaudhury, \textit{Senior Member, IEEE}, and Swapnil~D.~Dabhade 
\thanks{A part of this work was presented at IEEE ICIP 2015 \cite{Chaudhury2015}. This work was supported in part by the Startup Grant awarded by the Indian Institute of Science. Address: Department of Electrical Engineering, Indian Institute of Science, Bangalore 560012, India. Correspondence: kunal@ee.iisc.ernet.in.  
}}

\maketitle

\begin{abstract}
The bilateral filter is a non-linear filter that uses a range filter along with a spatial filter to perform edge-preserving smoothing of images.
A direct computation of the bilateral filter requires $O(S)$ operations per pixel, where $S$ is the size of the support of the spatial filter. 
In this paper, we present a fast and provably accurate algorithm for approximating the bilateral filter when the range kernel is Gaussian.
In particular, for box and Gaussian spatial filters, the proposed algorithm can cut down the complexity to $O(1)$ per pixel for any arbitrary $S$.
The algorithm has a simple implementation involving $N+1$ spatial filterings, where $N$ is the approximation order. 
We give a detailed analysis of the filtering accuracy that can be achieved by the  proposed approximation in relation to the target bilateral filter. This allows us to to estimate the order $N$ required to obtain a given accuracy. We also present comprehensive numerical results to demonstrate that the proposed algorithm is competitive with state-of-the-art methods in terms of speed and accuracy.
\end{abstract}


\IEEEpeerreviewmaketitle

\section{Introduction}

Gaussian and box filters typically work well in applications where the amount of smoothing required is small.
For example, they are quite effective in removing small dosages of noise from natural images. However, when the
noise floor is large, and one is required to average more pixels to suppress the noise, these filters begin 
to over-smooth sharp image features such as edges and corners. The over-smoothing can, however, be alleviated
using some form of data-driven (non-linear) diffusion, where the quantum of smoothing is controlled using the image features. 
A classical example in this regard is the famous PDE-based diffusion of Perona and Malik  \cite{Perona1990}. 
The bilateral filter was proposed by Tomasi and Maduchi \cite{Tomasi1998} as a filtering-based alternative to the
Perona-Malik diffusion. The bilateral filter has turned out to be a versatile tool that has found widespread applications in image processing, computer graphics, computer vision, and computational photography 
\cite{bilat_application_book}. More recently, the bilateral filter has received renewed attention in the context of image denoising \cite{Knaus2014,CR2015}. 

In this paper, we consider a standard form of the bilateral filter where a Gaussian kernel is used for range filtering, and a box or Gaussian kernel is used for spatial filtering \cite{Tomasi1998}. In this setting, the bilateral filtering of an image $\{ f(\i) : \i \in I\}$, where $I$ is some finite rectangular domain of $\Z^2$, is given by 
\begin{equation}
\label{BF}
 f_{\mathrm{BF}}(\i)=  \frac{\sum_{\j \in \Omega} w(\j) \  g_{\sigma_r}(f(\i-\j)-f(\i)) \ f(\i-\j)}{\sum_{\j \in \Omega} w(\j)  \  g_{\sigma_r}(f(\i-\j)-f(\i)) } 
\end{equation}
where 
\begin{equation}
\label{range_kernel}
\quad  g_{\sigma_r}(t) = \exp\left(- \frac{t^2}{2\sigma_r^2}\right).
\end{equation}
The spatial filter is a Gaussian:
\begin{equation}
\label{spatial_kernel1}
w(\i) = \exp\left(- \frac{\lVert \i \rVert^2}{2\sigma_s^2}\right) \qquad  (\i \in \Omega),
\end{equation}
or a box:
\begin{equation}
\label{spatial_kernel2}
w(\i) = 1/|\Omega|  \qquad (\i \in \Omega).
\end{equation}
The domain $\Omega$ of the spatial kernel  is a square neighbourhood, $\Omega=[-W,W] \times [-W,W]$, where $W=3\sigma_s$ for the Gaussian filter. 
We refer the interested reader to \cite{Tomasi1998,bilat_application_book} for a detailed exposition on the working of the filter. We note that the bilateral filter has a straightforward extension to video and volume data. Another natural extension is the cross (or joint) bilateral filter \cite{bilat_application_book}. While we will limit our discussion to the standard bilateral filter, the main ideas in this paper can also be applied to the above-mentioned extensions.

\subsection{Fast Bilateral Filtering}

It is clear that a direct computation of \eqref{BF} requires $O(W^2)$ operations per pixel. In fact, the computation is slow for practical settings of $W$. 
To address this issue, researchers have come up with several fast algorithms \cite{Durand2002} - \cite{Chaudhury2013}. Most of these are based on some form of approximation, and provide various levels of compromise between speed and quality of approximation. One of the early algorithms for fast bilateral filtering involved the quantization of the image intensities, where the final output was obtained via the interpolation of the output of a set of linear filters \cite{Durand2002}. It was later shown that this approximation can be used to obtain a constant-time implementation which further improves its speed \cite{Yang2009}. In a different direction, it was observed in \cite{Paris2006} that the bilateral filter can be conceived as a linear filter acting in three-dimensions, where the three-dimensions are obtained by augmenting the image intensity to the spatial dimensions.  This observation was used to derive a fast filtering in three-dimensions, which was then sampled to obtained the final output.  We refer the interested reader to \cite{Kamata2015} for a survey of fast algorithms for bilateral filtering.

The algorithms in \cite{Kamata2015,Porikli2008,Chaudhury2011} are particularly relevant to the present work. Here the authors  proceed by approximating \eqref{range_kernel} using polynomial and trigonometric functions, and demonstrate how the bilateral filter can be decomposed into a series of spatial filterings as result.  As is well-known, since spatial box and Gaussian filters can be implemented in constant-time using separability and recursion \cite{Deriche1993}, the overall approximation can therefore be computed in constant-time.  

\subsection{Present Contribution}

We propose a fast algorithm for computing \eqref{BF} which was motivated by the line of work in \cite{Chaudhury2011,Chaudhury2013}.
In particular, similar to these papers, we present a novel approximation of \eqref{range_kernel} that allows us to decompose the bilateral filter into a series of spatial convolutions.
The fundamental difference between the above papers and the present approach is that, instead of approximating \eqref{range_kernel} and then translating the approximation in range space, we directly approximate the translated Gaussians appearing in \eqref{BF}. In particular, the computational advantages obtained using the proposed approximation are the following: \\
\indent (1) For a fixed approximation order (to be defined shortly), the proposed approximation requires half the number of spatial filterings required by the approximations in \cite{Yang2009,Kamata2015,Chaudhury2011}. \\
\indent (2) The proposed approximation does not involve the transcendental functions $\cos( \omega x)$ and $\sin( \omega x)$ which are used in  \cite{Chaudhury2011,Chaudhury2013}. It only involves polynomials (and just a single Gaussian), and hence can be efficiently implemented on hardware \cite{Muller2006}. Moreover, the rounding error is small when working with polynomials.

As will be demonstrated shortly,  the proposed algorithm is generally faster and more accurate than Yang's algorithm \cite{Yang2009}, which is currently considered to be the state-of-the-art \cite{Kamata2015,errbilat}. In particular, we perform an error analysis whereby we compare the output obtained using the proposed algorithm with that of the exact bilateral filter. Due to the particular nature of the proposed approximation, our analysis is much more simple than that carried out for Yang's algorithm in \cite{errbilat}. Nevertheless, compared to Yang's algorithm, we are able to establish a smaller bound on the number of spatial filterings required to achieve a given filtering accuracy. The latter is defined in terms of the error between the outputs of the bilateral filter and the fast algorithm (this will be made precise in Section \ref{fastalgo}). 
To best of our knowledge, with the exception of \cite{Yang2009}, this is the only fast algorithm that comes with a provable guarantee on the quality of approximation.
At this point, we note that the term ``accurate''  is used in the paper not just to signify that the output of the fast algorithm is visibly close to that of the target bilateral filter.  
It also has a precise technical meaning, namely, that we can control the approximation order to make the error between the outputs of the bilateral filter and the fast algorithm arbitrarily small.

\subsection{Organization}

The rest of the paper is organized as follows. We present the proposed kernel approximation and the error analysis in Section \ref{sec1}. In Section \ref{fastalgo}, we develop a fast constant-time algorithm arising from the Gaussian-polynomial approximation. We then analyze the quality of approximation that can be achieved using our algorithm. This gives us a simple rule for tuning the approximation order for a given user-defined accuracy. We present exhaustive numerical results in Section \ref{exp}, and demonstrate the superior performance of the proposed algorithm over some of the existing algorithms.

\section{Gaussian-Polynomial Approximation}
\label{sec1}

The present idea is to consider the translated kernel $g_{\sigma_r}(t - \tau)$ that appears in \eqref{BF}, where $t=f(\i-\j)$ and $\tau=f(\i)$. We can write 
\begin{equation}
\label{decomp}
g_{\sigma_r}(t - \tau) = \exp\left(- \frac{\tau^2}{2\sigma_r^2}\right) \exp\left(- \frac{t^2}{2\sigma_r^2}\right) \exp\left(\frac{\tau t}{\sigma_r^2}\right).
\end{equation}
For a fixed translation $\tau$, this is a function of $t$. Notice that the first term is simply a scaling factor, while the second term is a Gaussian centered at the origin. In fact, the second term essentially contributes to the bell shape of the translated Gaussian. The third term is a monotonic exponential, which is increasing or decreasing depending on the sign of $\tau$; this term helps in translating the Gaussian to $t=\tau$. 

We assume (without loss of generality, as will be explained at the start of Section \ref{fastalgo}) that the dynamic range of the image is $[-T,T]$. That is, the arguments $t=f(\i-\j)$ and $\tau=f(\i)$ in \eqref{decomp} take values in $[-T,T]$. This means that the product $ \tau t$ appearing in \eqref{decomp} takes values in $[-T^2,T^2]$. Consider the Taylor expansion of the exponential term about the origin:
\begin{equation}
\label{approx}
\exp\left(\frac{\tau t}{\sigma_r^2}\right) = \sum_{n=0}^{N-1} \frac{1}{n!}  \left(\frac{\tau t}{\sigma_r^2}\right)^n + \text{ higher-order terms}.
\end{equation}
By dropping the higher-order terms, we obtain the following approximation of \eqref{decomp}:
\begin{equation}
\label{GaussPolynomial}
\phi_{N,\sigma_r}(t,\tau)= \exp\left(- \frac{t^2+\tau^2}{2\sigma_r^2}\right)  \Bigg[\sum_{n=0}^{N-1} \frac{1}{n!}  \left(\frac{\tau t}{\sigma_r^2}\right)^n \Bigg].
\end{equation}
Being the product of a bivariate Gaussian and a polynomial, we will henceforth refer to \eqref{GaussPolynomial} as a \textit{Gaussian-polynomial}, where $N$ is its approximation \textit{order}. By construction, we have the pointwise convergence
\begin{equation}
\label{asymp2}
\lim_{N \rightarrow \infty}  \phi_{N,\sigma_r}(t, \tau) = g_{\sigma_r}(t - \tau).
\end{equation}
We would like to note that the above idea of splitting the kernel and approximating a part of its using Taylor polynomials was employed in \cite{YDG2003} in the context of the fast Gauss transform. To the best of our knowledge, this idea has not been exploited for fast bilateral filtering along the lines of the present work.

\begin{figure}
\centering
\includegraphics[width=0.6\linewidth]{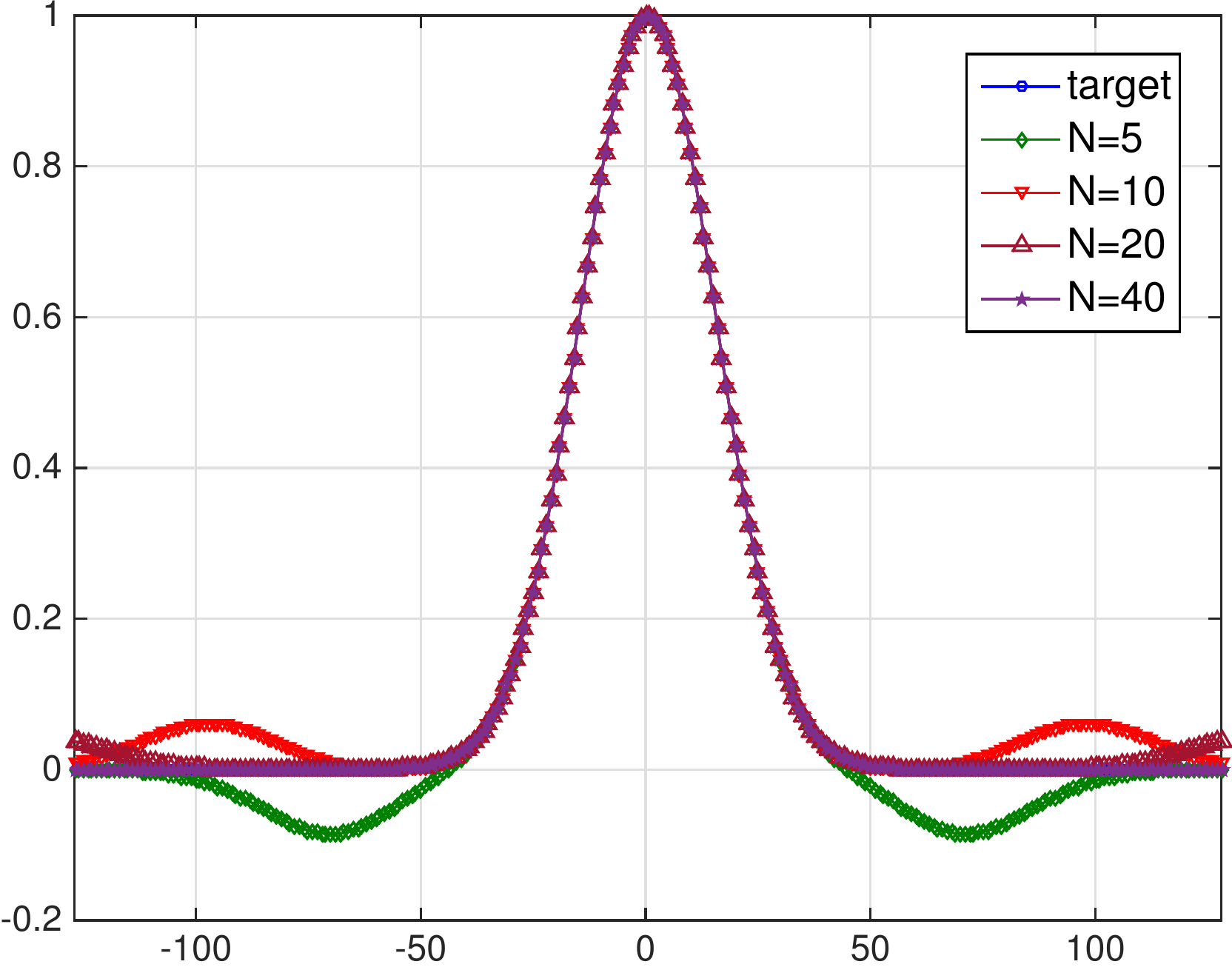}
\caption{Approximation of $g_{30}(t - \tau)$ using Gaussian-polynomials $\phi_{N,30}(t, \tau)$ with different $N$. The bivariate functions $g_{30}(t - \tau)$ and $\phi_{N,30}(t, \tau)$ have been sampled along  $t=-\tau$ to generate a one-dimensional profile.} 
\label{oneDimScan}
\end{figure}

\begin{figure}
\centering
\subfloat[]{\includegraphics[width=0.5\linewidth]{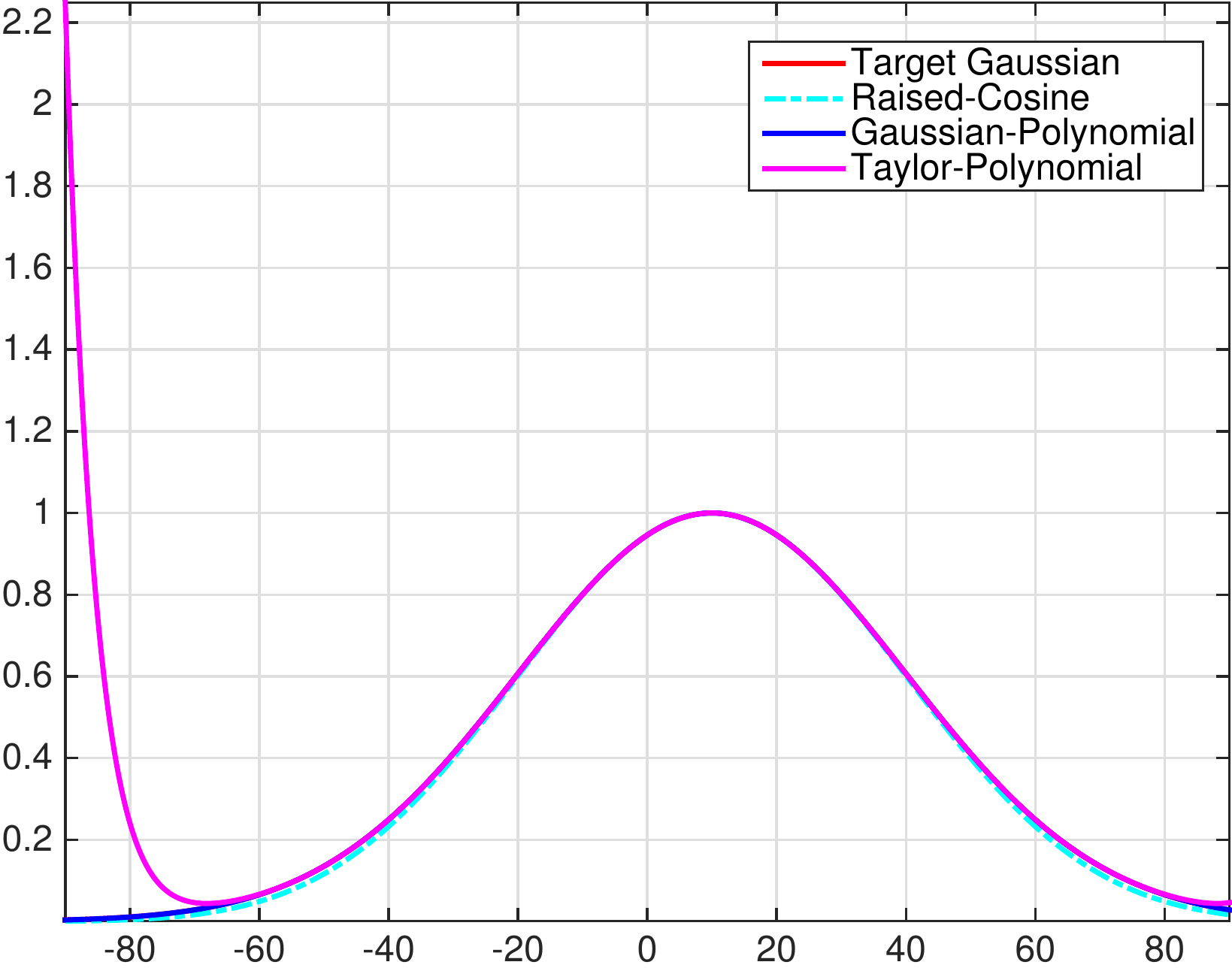}}  
\subfloat[]{\label{fb}\includegraphics[width=0.5\linewidth]{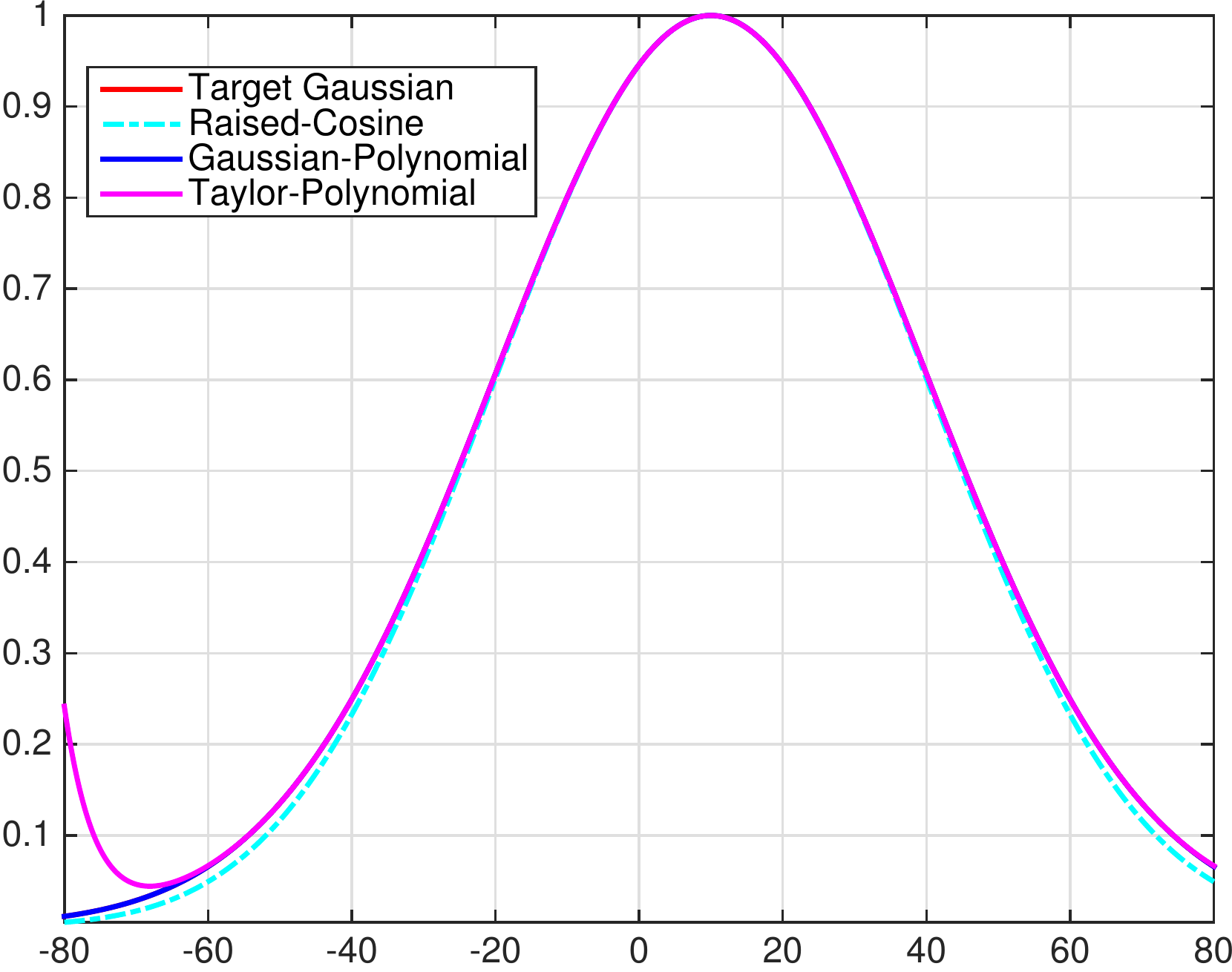}}  
\caption{Comparison of the approximations of  $g_{30}(t - 10)$ using raised-cosine \cite{Chaudhury2011}, Taylor polynomial \cite{Porikli2008}, and Gaussian-polynomial of order $10$. We notice in (a) that the Taylor polynomial quickly goes off to $+\infty$ as one moves away from the origin. For this reason, we restricted the plot to $[-90,90]$, although the desired approximation range is the full dynamic range $[-128,128]$. The plots over $[-80,80]$ are separately provided in (b) for comparing the raised-cosine and the Gaussian-polynomial approximations with the target Gaussian.} 
\label{comparison}
\end{figure}

In Figure \ref{oneDimScan}, we study the approximations corresponding to different $N$. The fundamental difference between the Taylor approximation in \cite{Porikli2008}  and  the Gaussian-polynomial approximation \eqref{asymp2} is that instead of approximating the entire Gaussian, we approximate one of its component, namely the exponential function in \eqref{decomp}. The intuition behind this is that the Taylor polynomial blows up as one moves away from the origin. This makes it difficult to approximate the tail part of a Gaussian using such polynomials. On the other hand, the exponential in \eqref{decomp} is monotonic,  and hence can be closely approximated using polynomials. This point is explained with an example in Figure \ref{comparison}. In particular, notice in Figure \ref{fb} that the Gaussian-polynomial approximation is quite precise over the range of interest, and is comparable to the raised-cosine approximation of same order \cite{Chaudhury2011}.

\subsection{Quantitative Error Analysis}

Before explaining how we can use Gaussian-polynomials to derive a fast bilateral filter in Section \ref{fastalgo}, we study the kernel error incurred by approximating \eqref{range_kernel} using Gaussian-polynomials. We will see in Section \ref{fastalgo} that a bound on the kernel error can in turn be used to bound the filtering accuracy of the fast algorithm. Note that \eqref{asymp2} tells us that Gaussian-polynomial can be used to approximate the range kernel with arbitrary accuracy. However, in practice, we will be required to use a Gaussian-polynomial of some fixed order $N$. A relevant question is the size of error incurred for a given $N$? A related question is that, given some error margin $\varepsilon >0$, how do we fix the smallest $N$ such that the corresponding error is within $\varepsilon$? 

To begin with, we define the error function
\begin{align}
\label{error}
E_{N,\sigma_r}(t,\tau)&=g_{\sigma_{r}}(t-\tau) - \phi_{N,\sigma_r}(t,\tau) \nonumber \\
& = \exp\left(- \frac{t^2+\tau^2}{2\sigma_r^2} \right)  \sum_{n=N}^{\infty} \frac{1}{n!}  \left(\frac{\tau t}{\sigma_r^2}\right)^n.
\end{align}
The mathematical problem is one of bounding \eqref{error} for fixed $N$ and $\sigma_r$.  In this work, we consider the $\ell_{\infty}$ error given by 
\begin{equation}
\label{Linf}
\lVert  E_{N,\sigma_r} \rVert_{\infty} = \max \Big\{ |E_{N,\sigma_r}(t,\tau)| : \ -T \leq t,\tau \leq T \Big\}.
\end{equation}
This is also referred to as the worst-case or uniform error. We note that one can measure the error using other means, e.g., using the $\ell_2$ metric. The reason why we choose the $\ell_{\infty}$ metric is that our ultimate goal is to quantify the $\ell_{\infty}$ accuracy of the final filtering arising from the approximation, and a bound on \eqref{Linf} is sufficient for this purpose. Moreover, computing the $\ell_{\infty}$ error is relatively simple.

Using the inequality $(t^2+\tau^2)/2 \geq |\tau t|$, we can bound the first term in \eqref{error} by $\exp(-|\tau t| / \sigma_r^2)$. Therefore, we have
\begin{equation}
\label{trivial_bound}
\lVert  E_{N,\sigma_r} \rVert_{\infty} \leq \max_{s \in [0,T^2]}  \ \psi_{N,\sigma_r}(s),
\end{equation}
where 
\begin{equation}
\label{psi}
\psi_{N,\sigma_r}(s) = \exp\left(- \frac{s}{\sigma_r^2} \right)  \Bigg[\sum_{n=N}^{\infty} \frac{1}{n!} \left(\frac{s}{\sigma_r^2}\right)^n \Bigg].
\end{equation}
\begin{figure}
\centering
\subfloat[]{\includegraphics[width=0.5\linewidth]{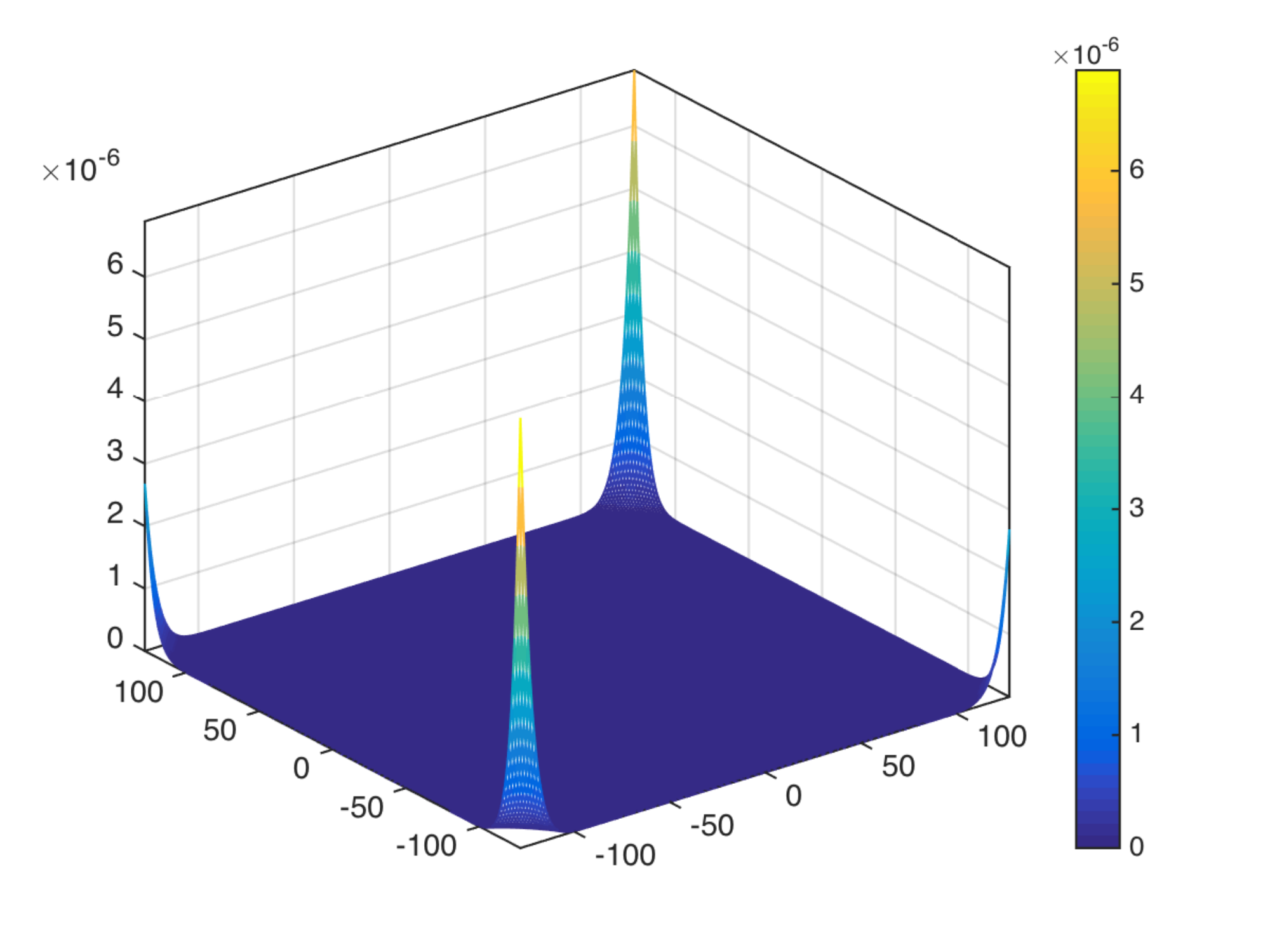}}  
\subfloat[]{\includegraphics[width=0.5\linewidth]{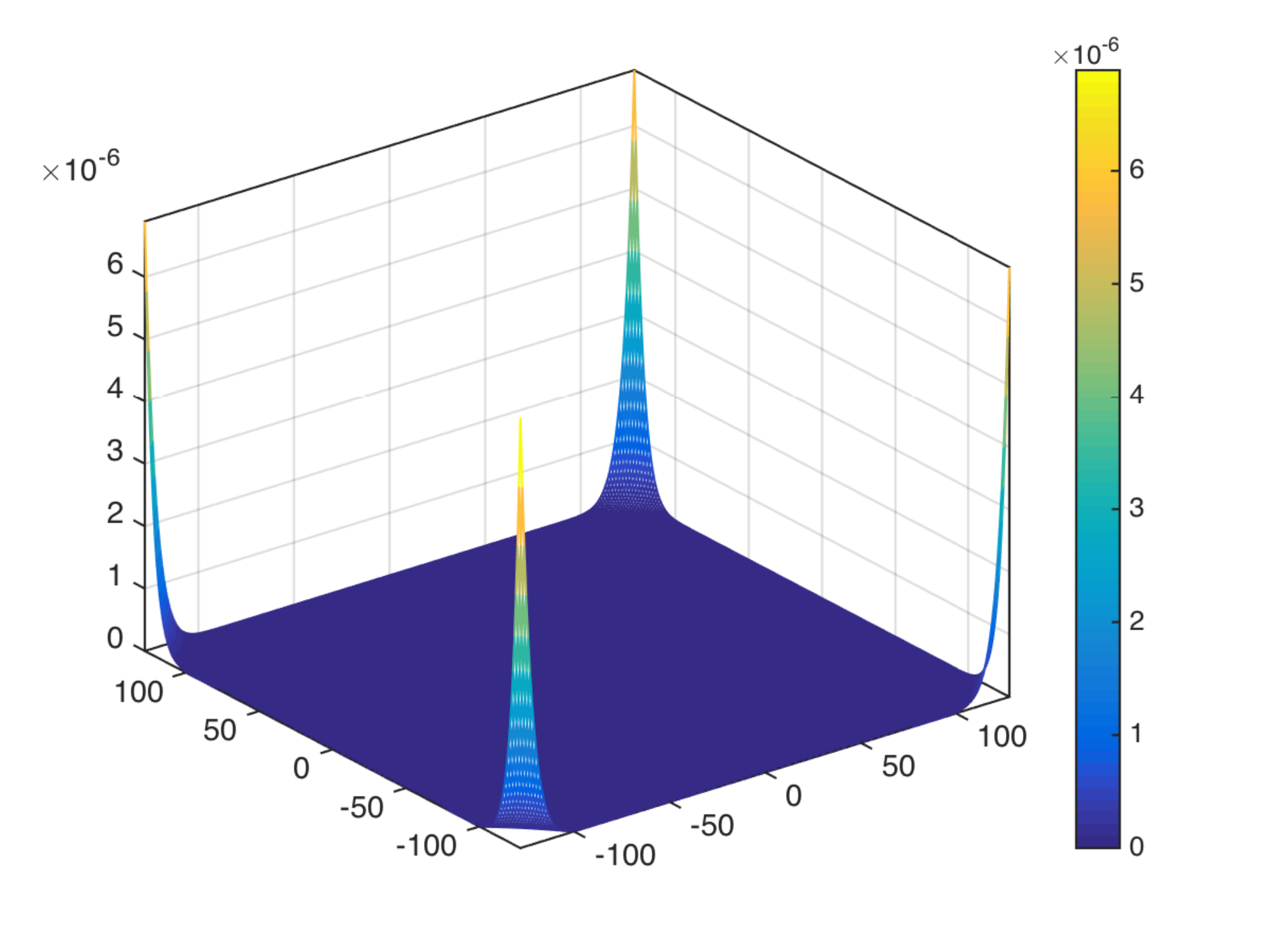}}  
\caption{Comparison of the actual error \eqref{error} and the bound in \eqref{psi} for $T = 128$ and $\sigma_r=30$. We plot the samples of the error function $E_{40,30}(t,\tau)$ over the square domain $-128 \leq t,\tau \leq 128$ in (a). We compare this with the samples of \eqref{psi} over the same domain in (b), where we have set $s=|\tau t|$. Notice that the supremum of either plots are of the same order of magnitude.}  
\label{errorBound}
\end{figure}
Using \eqref{trivial_bound}, we obtain the following result. We note that this bound is stronger than that derived for the fast Gauss transform in \cite{YDG2003}.
\begin{proposition} 
\label{prop_tightBound}
\begin{equation}
\label{upperBound}
\lVert  E_{N,\sigma_r} \rVert_{\infty} \leq  \sum_{n=N}^{\infty} \frac{e^{-\lambda}  \lambda^n }{n!}  \qquad \big(\lambda = T^2/\sigma_r^2\big).
\end{equation}
\end{proposition}
To arrive at \eqref{upperBound}, we proceed by writing \eqref{psi} as
\begin{equation*}
\psi_{N,\sigma_r}(s) = 1-  \exp\left(- \frac{s}{\sigma_r^2} \right)  \sum_{n=0}^{N-1} \frac{1}{n!}  \left(\frac{s}{\sigma_r^2}\right)^n .
\end{equation*}
After differentiation, we get
\begin{equation*}
\psi'_{N,\sigma_r}(s) = \frac{1}{ (N-1)! \sigma_r^2} \left(\frac{s}{\sigma_r^2}\right)^{N-1} \exp\left(- \frac{s}{\sigma_r^2} \right)  \geq 0 .
\end{equation*}
Thus, \eqref{psi} is non-decreasing on $[0,T^2]$, whereby we conclude that the maximum in \eqref{trivial_bound} is attained at $s=T^2$. This establishes Proposition \ref{prop_tightBound}.

To get an idea of the tightness of the bound in \eqref{upperBound}, we compare the mesh plots of \eqref{error} and \eqref{psi} in Figure \ref{errorBound} when $\sigma_r=30$ and $N=40$. While there is a gap between the error and the corresponding bound at certain values of $(t,\tau)$, the supremum of the latter (which occurs at one of the boundaries as predicted above) is nevertheless of the same order of magnitude as the supremum of the former. 

\subsection{Relation between $N$ and Kernel Error}

Having obtained a bound on the approximation error, we consider the  problem of finding the smallest $N$ such that \eqref{Linf} is within some allowed error margin $\varepsilon>0$.  
Note that the quantity on the right in \eqref{upperBound} is simply the tail probability of a Poisson random variable with parameter $\lambda$. We recall that a random 
variable $X$ taking values in $\{0,1,2,\ldots\}$ is said to follow a Poisson distribution with parameter $\lambda > 0$ if  
\begin{equation*}
\mathrm{Prob}(X=n) = \frac{e^{-\lambda}  \lambda^n }{n!} \qquad (n=0,1,2,\ldots).
\end{equation*}
We can thus interpret the quantity on the right in \eqref{upperBound} as the probability $\mathrm{Prob}(X \geq N)$. In this context, the leading question is the following: given $\varepsilon>0$, find the smallest $N$ such that $\mathrm{Prob}(X \geq N) \leq \varepsilon$.
The advantage of expressing the problem in this form is that it brings to our disposal various tools for bounding the tail probability. For example, assuming that $N > \lambda$, we have the Chebyshev bound \cite{prob}:
\begin{equation}
\label{cheby1}
 \mathrm{Prob}(X \geq N) \leq \frac{\lambda}{(N-\lambda)^2}.
\end{equation}
On the other hand, the Chernoff bound \cite{prob} when $N > \lambda$ is given by
\begin{equation}
\label{Chernoff}
 \mathrm{Prob}(X \geq N) \leq  \frac{e^{-\lambda}{(e \lambda)}^N}{N^N}.
\end{equation}
Numerical experiments suggest that for $\sigma_r < 70$ and for a range of values of $\varepsilon$ (to be reported shortly), the empirically computed $N$ is always larger than $\lambda$. Under this assumption, we have the following estimate of the smallest $N$ using \eqref{cheby1}:
\begin{equation}
\label{cheby2}
N_0 = [\lambda + \sqrt{ \lambda/\varepsilon}],
\end{equation}
where $[x]$ is the smallest integer greater than or equal to $x$. 

\begin{table}
\caption{Comparison of the Gaussian-polynomial order obtained using \eqref{LambW}, where (1) $W_0$ is computed using the Matlab function \texttt{lambertw} ($N'_0$),  (2) $W_0$ is given by \eqref{series} ($N_0''$), and (3) the series evaluation is refined using three Newton iterations ($N_0'''$).}
\centering
\begin{tabular}{ |p{0.3cm} ||p{0.4cm}|p{0.4cm}|p{0.4cm}|p{0.4cm}|p{0.4cm}|p{0.4cm}|p{0.4cm}|p{0.4cm}|p{0.4cm}| }
 \hline
$\sigma_r$       & 10  & 15  & 20 & 25 & 30 & 35 & 40 & 45 & 50 \\
\hline
$N'_0$   & 214 & 107 & 67 & 48 & 37 & 30 & 25 & 21 & 19 \\
\hline
$N''_0$ & 270 & 124 & 74 & 50 & 37 & 30 & 25 & 21 & 19 \\
\hline
$N'''_0$    & 214 & 107 & 67 & 48 & 37 & 30 & 25 & 21 & 19\\
\hline
\end{tabular}
\label{table1}
\end{table} 

As is well-known, the Chernoff bound \eqref{Chernoff} is typically tighter than the Chebyshev bound. However, finding the smallest $N$ such that
\begin{equation}
\label{inequality}
 \frac{e^{-\lambda}{(e \lambda)}^N}{N^N} \leq \varepsilon
\end{equation}
is somewhat more involved. 

\begin{proposition} Let  $t \mapsto W_0(t)$ be the inverse of the map $t \mapsto t \exp(t)$ on $(0,\infty]$. Then the smallest integer greater  than $\lambda$ for which \eqref{inequality} holds is
\begin{equation}
\label{LambW}
N_0=[q/W_0(q e^{-p})],
\end{equation}
where $p=1+\log(\lambda)$ and $q=-\lambda-\log \varepsilon$.
\end{proposition}
The details are provided in Appendix \ref{proof1}. While $W_0(t)$ can be computed using the Matlab script \texttt{lambertw(0,t)}, we note that $W_0(t)$ can be approximated using a series expansion \cite{lamb}. In particular, the first four terms are
\begin{equation}
\label{series}
 W_0(t) = t -t^2+ \frac{3}{2} t^3 - \frac{8}{3} t^4.
\end{equation}
However, we observed that \eqref{series} provides inexact estimates when $\lambda$ is large, that is, when $\sigma_r$ is small. An extremely large number of terms of the series are required to get a precise estimate. To address this problem, we propose to use Newton iterations for finding the positive root of $\nu(x)=x \log x- px - q = 0$ (see Appendix \ref{proof1} for notations), where the initialization is done using \eqref{LambW} and \eqref{series}. Namely, starting with $x_0=q/W_0(q e^{-p})$, we run the following iterations for $k \geq 0$:
\begin{align}
\label{Newton_method} 
x_{k+1}&= x_k - \frac{\nu(x_k)}{\nu'(x_k)} = x_k - \frac{x_k \log x_k - px_k-q }{\log x_k+1-p}.
\end{align}
In practice, we noticed that about $3$-$4$ iterations are sufficient to produce a good solution. In Table \ref{table1}, we illustrate the improvement obtained after performing the Newton iterations. The complete scheme for computing the order  for a given accuracy $\varepsilon$ is summarized in Algorithm \ref{algo1}. 
\IncMargin{2mm}
\begin{algorithm}
\KwData{$\sigma_r,\varepsilon,T$.}
\KwResult{$N_0$.}
\eIf{$\sigma_r \geq 70$}{$N_0=10$\;}
{
$\lambda = (T/\sigma_r)^2$\;
$p=1+\log(\lambda)$\;
$q = - \lambda - \log \varepsilon$\;
$t = q/(e\lambda)$\;
$W_0 = t - t^2 + 3t^3/2 - 8t^3/4$\;
$N_0 = q/W_0$\;
\If{$\sigma_r < 30$}{
\For{$k=1,2,3$}{
    $N_0=N_0- \frac{N_0 \log(N_0)-pN_0-q}{\log N_0+1-p}$\;
    }
}
}
$N_0=[N_0]$;
\caption{\small Estimation of the approximation order.}
\label{algo1}
\end{algorithm}
\DecMargin{1.5mm}

\begin{figure}
\centering
\subfloat[$\varepsilon=0.1$.] {\label{}\includegraphics[width=0.5\linewidth]{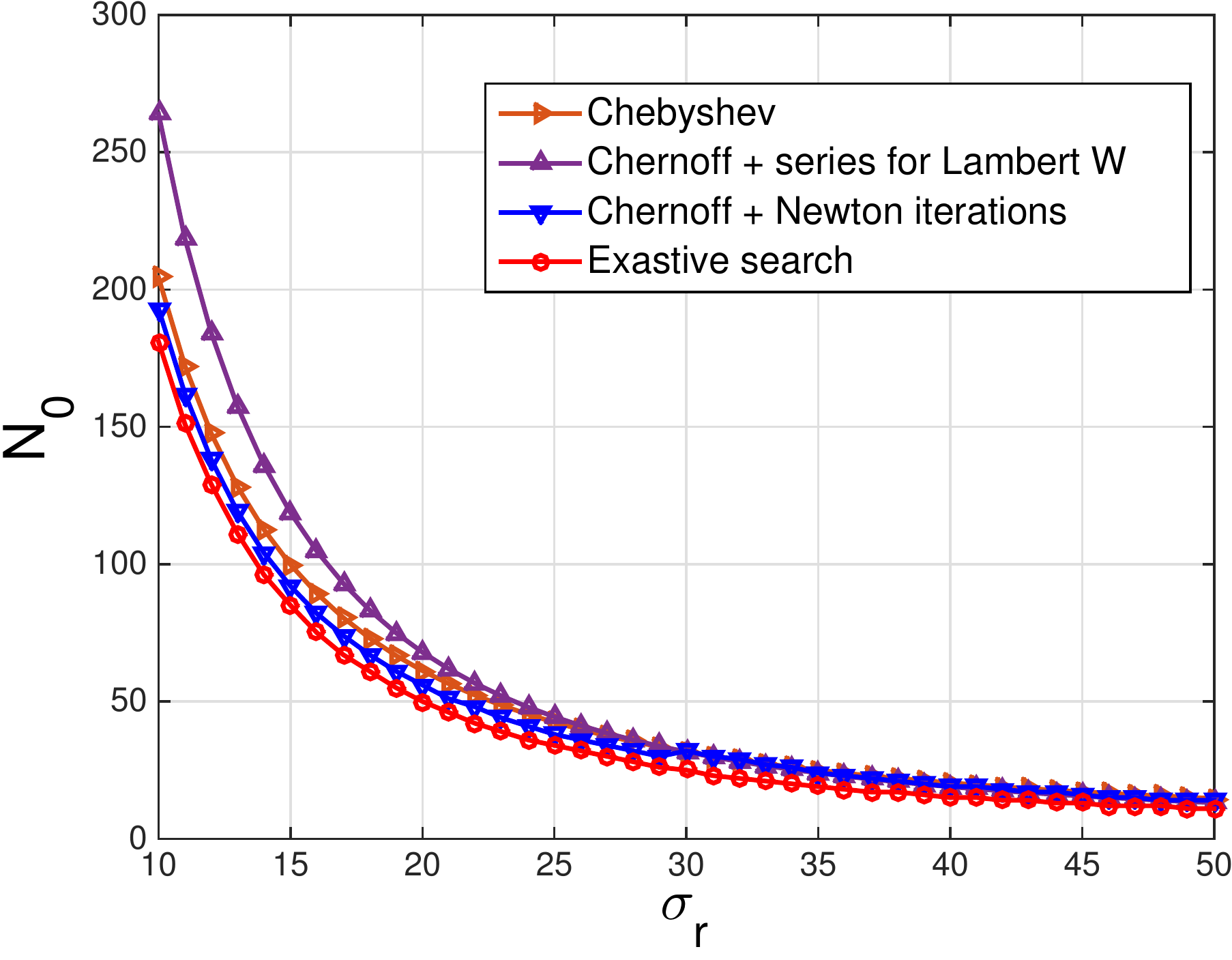}}  
\subfloat[$\varepsilon=0.001$. ] {\label{}\includegraphics[width=0.5\linewidth]{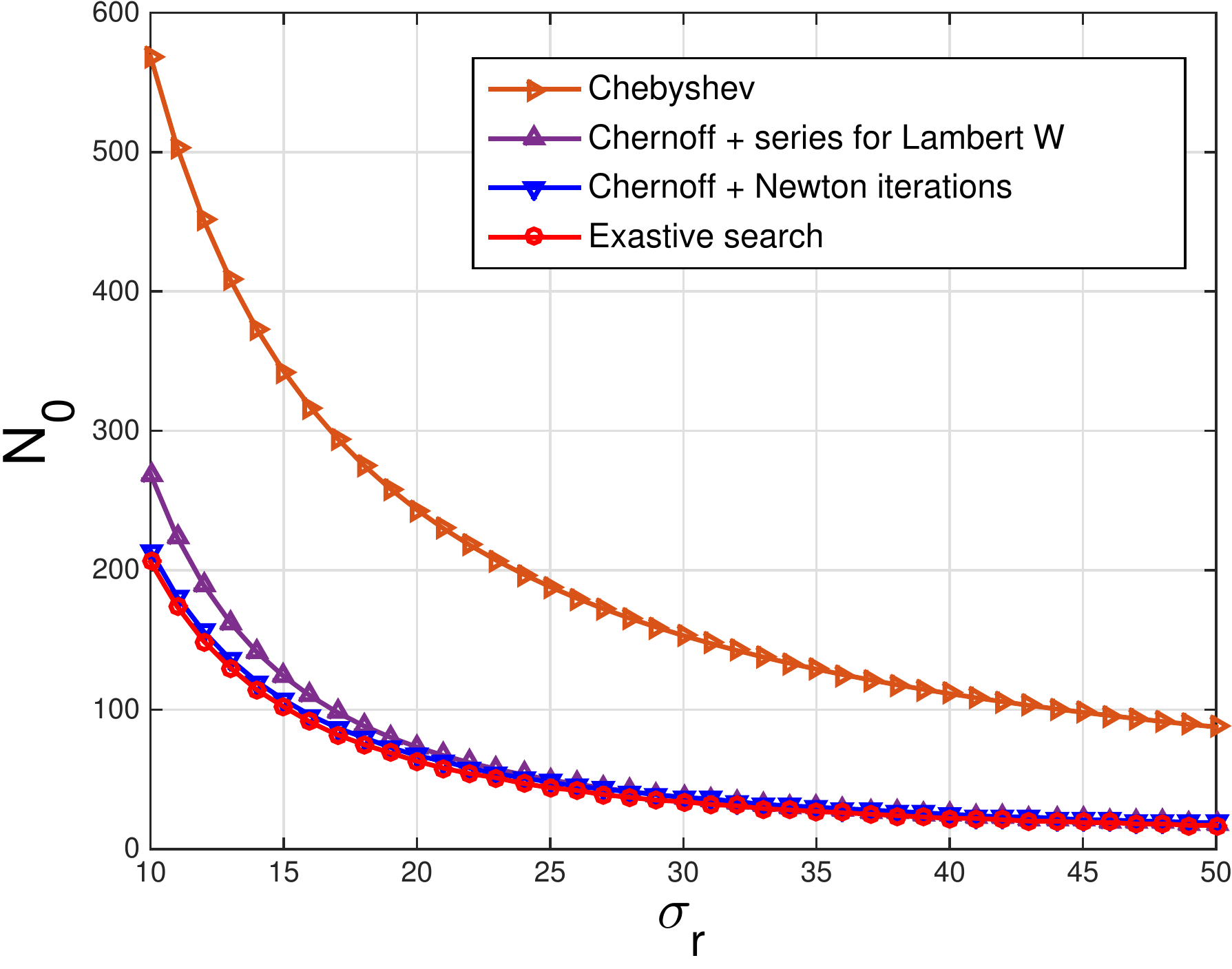}} \\
\subfloat[$\varepsilon=0.1$. ] {\label{}\includegraphics[width=0.5\linewidth]{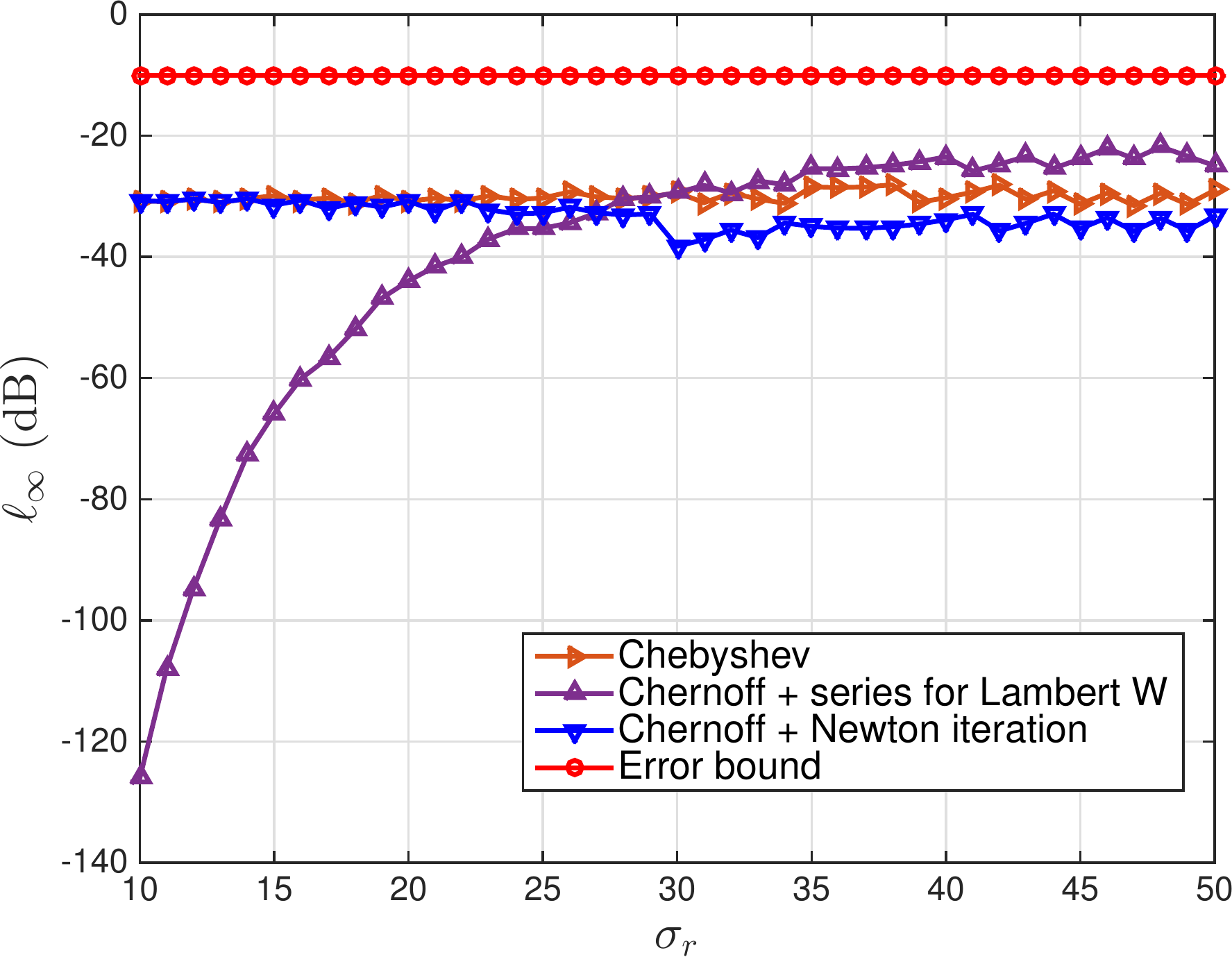}}  
\subfloat[$\varepsilon=0.001$. ] {\label{}\includegraphics[width=0.5\linewidth]{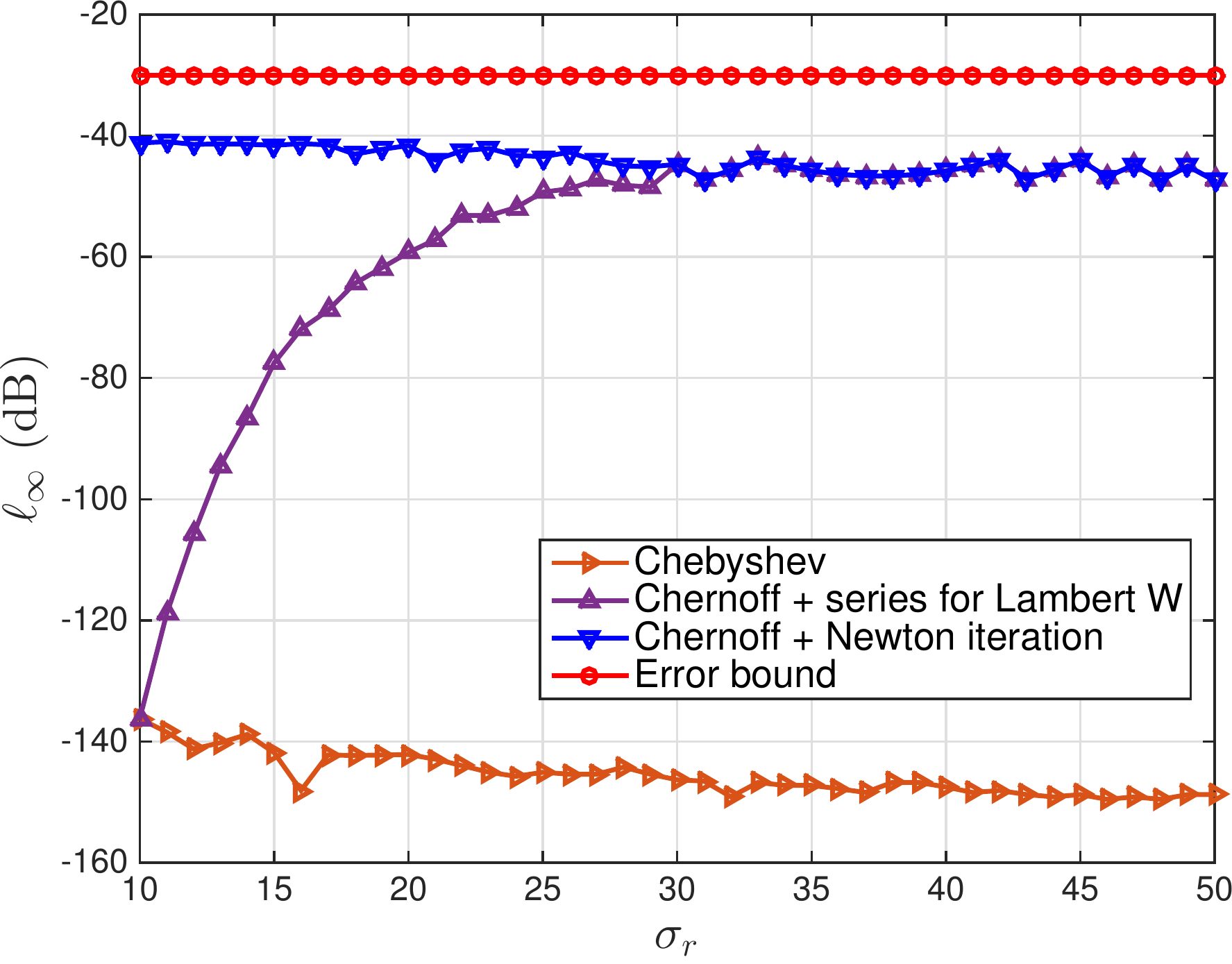}}  
\caption{For $\varepsilon=0.1$ and $0.001$, we compare the order $N_0$ obtained using various methods (top row) and the corresponding error (bottom row).}
\label{tightness}
\end{figure}

Note that for $\sigma_r > 70$, we use a fixed order of $10$. This is because the condition $N > \lambda$ in \eqref{cheby1} and \eqref{Chernoff} is violated in this regime. 
Moreover, we have noticed that a small order suffices when $\sigma_r$ is large.
In Figure \ref{tightness}, we compare the estimated order $N_0$ obtained using the following methods: Chebyshev \eqref{cheby2}, Chernoff \eqref{LambW} along with \eqref{series}, and Chernoff followed by Newton iterations \eqref{Newton_method}. We also compare the corresponding errors  (computed using exhaustive search) given by \eqref{Linf}. 
Notice that the estimates are close to that obtained using exhaustive search when $\varepsilon=0.1$; however, when $\varepsilon=0.001$, the Chebyshev bound is quite loose.

\section{Fast Bilateral Filtering}
\label{fastalgo}

We now explain how Gaussian-polynomials can be used to derive a fast algorithm for implementing \eqref{BF}. As a first step, we center the intensity range 
$\{ f(\i) : \i \in I\}$ around the origin. This is in keeping with the Taylor expansion in \eqref{GaussPolynomial} which is performed around the origin. 
A simple means of doing so is to set $t_c=T$, assuming the dynamic range to be $[0,2T]$, and to consider the centred image $\{h(\i) : \i \in I\}$ given by
\begin{equation}
\label{shift}
h(\i) = f(\i) - t_c \qquad (\i \in I).
\end{equation}
The crucial observation is that that the shift operation in \eqref{shift} commutes with the non-linear bilateral filtering.

\begin{proposition} For $\i \in I$,
\begin{equation}
\label{centering}
f_{\mathrm{BF}}(\i) = h_{\mathrm{BF}}(\i) + t_c. 
\end{equation}
\end{proposition}
In other words, we can first centre the intensity range, apply the bilateral filter, and finally add back the centre to the output.
Henceforth, we will assume that the range of the input image is $[-T,T]$. For an $8$-bit grayscale image, $T=128$.  

\subsection{Fast Algorithm}

The underlying mechanism of the proposed fast algorithm is related to the fast algorithms in \cite{Chaudhury2011,Chaudhury2013}. The subtle difference is that instead of directly approximating \eqref{range_kernel}, we approximate its translates in \eqref{BF}. In particular, we fix some order $N$, and approximate the range kernel in \eqref{BF} using \eqref{GaussPolynomial}. This gives us the following Gaussian-Polynomial Approximation (GPA) of \eqref{BF}:
\begin{equation}
\label{GPA}
\! \ f_{\mathrm{GPA}}(\i)=  \frac{\sum_{\j \in \Omega}\! w(\j)   \phi_{N,\sigma_r}(f(\i-\j)-f(\i))  f(\i-\j)}{\sum_{\j \in \Omega} w(\j)    \phi_{N,\sigma_r}(f(\i-\j)-f(\i))}.
\end{equation}
Next, for $n=0,\ldots,N-1$, we define the images 
\begin{equation}
\label{intImg}
G_n(\i) =  \left(\frac{f(\i)}{\sigma_r}\right)^n \!,  \ \ F_n(\i) = \exp\left(- \frac{f(\i)^2}{2\sigma_r^2}\right) G_n(\i),
\end{equation}
and set
\begin{equation}
\label{spatialFiltering}
\bar{F}_n(\i) = \left(F_n \ast w \right)(\i) = \sum_{\j \in \Omega} w(\j) F_n(\i - \j).
\end{equation}
We can then write \eqref{GPA} as (cf. Appendix \ref{fastAlgo})
\begin{equation}
\label{PbyQ}
f_{\mathrm{GPA}}(\i)= \frac{P(\i)}{Q(\i)},
\end{equation}
where
\begin{equation}
\label{P}
P(\i) = \sigma_r  \sum_{n=0}^{N-1} \frac{1}{n!} G_n(\i)  \bar{F}_{n+1}(\i),
\end{equation}
and
\begin{equation}
\label{Q}
Q(\i) =  \sum_{n=0}^{N-1} \frac{1}{n!} G_n(\i) \bar{F}_{n}(\i).
\end{equation}

Notice that we have effectively transferred the non-linearity of the bilateral filter to the intermediate images in \eqref{intImg}, which are obtained from the input image using simple pointwise transforms. The computational advantage that we get from the above manipulation is that the spatial filtering in \eqref{spatialFiltering} can be computed using $O(1)$ operations per pixel when $w$ is a box or a Gaussian \cite{Deriche1993}. The overall cost of computing \eqref{GPA} is therefore $O(1)$ per pixel with respect to the filter size $W$. This is a substantial reduction from the $O(W^2)$ complexity of the direct implementation of \eqref{BF}.

The complete algorithm for computing \eqref{GPA} is summarized in Algorithm \ref{GPAcode}, which we will continue to refer as GPA. Note that we efficiently implement steps \eqref{intImg} to \eqref{Q} by avoiding redundant computations. In particular, we recursively compute the images in \eqref{intImg} and the factorials in \eqref{P} and \eqref{Q}. Notice that steps \ref{start1}-\ref{end1}, \ref{start2}-\ref{end2}, \ref{start3}-\ref{end3}, and \ref{final} are cheap pointwise operations. The main computation in Algorithm \ref{GPAcode} is the spatial filtering in step \ref{filt2}, and the initial filtering in step \ref{filt1}. That is, the overall cost is dominated by the cost of computing $N+1$ spatial filtering. In this regard, we note that for the same degree $N$, the number of  spatial filterings required in \cite{Chaudhury2011,Chaudhury2013} is $4N$, and that in \cite{Yang2009} is $2N$. Moreover, we note that the proposed algorithm involves the evaluation of a transcendental function just once, namely in step \ref{transc}. In contrast, the algorithm in \cite{Yang2009} requires $N$ evaluations of the Gaussian over the whole image. Thus, the present algorithm has smaller rounding errors, and is better suited for hardware implementations \cite{Muller2006} compared to the above mentioned algorithms. Yet another key advantage with the Algorithm \ref{GPAcode} is that we need just six images (excluding the input and output images) for the complete pipeline. As against this, the algorithm in \cite{Yang2009} requires the computation and storage of $N$ \textit{principal} images, which are interpolated to get the final output.

\IncMargin{2mm}
\begin{algorithm}
\textbf{Input}: $\{f(\i) : \i \in I\}$ taking values in $[0,2T]$\;
\textbf{Spatial Filter}: $\Omega$ and $\{w(\i) : \i \in \Omega\}$\; 
\textbf{Parameters}:  $\sigma_r$ and $N$\;
\textbf{Output}: $\{f_{\mathrm{GPA}}(\i) : \i \in I\}$ given by \eqref{GPA}\;
\For{$i \in I$}{
$h(\i) = f(\i) - T$\; \label{start1}
$F(\i) = \exp(-h(\i)^2 / 2\sigma_r^2)$\; \label{transc}
$G(\i) =1$\; 
$P(\i)=0$\;
$Q(\i)=0$\;
$H(\i)=h(\i)/\sigma_r$\; \label{end1}
}
$\bar{F}(\i) = \left(F \ast w \right) (\i)$\; \label{filt1}
\For{$n=1,\ldots,N$}{
\For{$i \in I$}{
$Q(\i)= Q(\i) +  G(\i) \bar{F}(\i)$\; \label{start2}
$F(\i) = H(\i)  F(\i)$\; \label{end2}
}
$\bar{F}(\i) = \left(F \ast w \right) (\i)$\; \label{filt2}
\For{$i \in I$}{
$P(\i) = P(\i) +  G(\i) \bar{F}(\i)$\; \label{start3}
$G(\i) = H(\i) G(\i)/n$\; \label{end3}
}
}
\For{$i \in I$}{
$f_{\mathrm{GPA}}(\i)= \sigma_r  \left(P(\i)/Q(\i)\right) + T$\; \label{final}
}
\caption{\small Gaussian-Polynomial Approximation (GPA).}
\label{GPAcode}
\end{algorithm}
\DecMargin{2mm}

\subsection{Filtering Accuracy}

It is clear that the kernel error, and hence the overall quality of approximation, is controlled by the order $N$. In this regard, we need a rule to fix $N$ in Algorithm \ref{GPAcode}.  
As before, we will consider the worst-case error given by
\begin{equation}
\label{err_infinity}
\lVert  f_{\mathrm{BF}} -  f_{\mathrm{GPA}} \rVert_{\infty} = \max \Big\{ |f_{\mathrm{BF}}(\i) -  f_{\mathrm{GPA}}(\i)| : \i \in I \Big\}.
\end{equation}
By bounding \eqref{err_infinity}, we can control the pixelwise difference between the exact and the approximate bilateral filter. 
In particular, we have the following result which formally establishes the intuitive fact that the filtering accuracy is essentially within a certain factor of the kernel error given by \eqref{Linf}. The details of the derivation are provided in Appendix \ref{proof2}.

\begin{proposition} Suppose that the spatial filter is non-negative and normalized, i.e., $w(\j) \geq 0$ for all $\j \in \Omega$, and
\begin{equation}
\label{normal}
\sum_{\j \in \Omega} w(\j)=1.
\end{equation}
 Then
\begin{equation}
\label{accuracy}
\lVert  f_{\mathrm{BF}} -  f_{\mathrm{GPA}} \rVert_{\infty}  \leq 2\frac{T \lVert E_{N,\sigma_r} \rVert_{\infty}  }{w(0) - \lVert  E_{N,\sigma_r}  \rVert_{\infty} }.
\end{equation}
\end{proposition}
We note that the spatial filters \eqref{spatial_kernel1} and \eqref{spatial_kernel2} are non-negative, and that $w(\i)$ appears in both the numerator and denominator of \eqref{BF} and \eqref{GPA}. Therefore, we can assume \eqref{normal} without any loss of generality. In fact, \eqref{normal} is automatically true for the box filter. We also recall that the range of the image is assumed to be centered; the intensity values are in the interval $[-T,T]$, where $T \approx 128$ for most grayscale images.

\subsection{Relation between Accuracy and $N_0$}

Note that by combining \eqref{accuracy} with the bound in  \eqref{Chernoff}, we get a direct control on the filtering accuracy in terms of the approximation order. In particular, suppose that we want \eqref{err_infinity} to be within $\pm \delta$. A sufficient condition for this is that
\begin{equation*}
\lVert  E_{N,\sigma_r}  \rVert_{\infty}  \leq \frac{w(0)  \delta}{2 T+\delta} .
\end{equation*}
To summarize, we have the following guarantee that follows from \eqref{accuracy}.
\begin{corollary} Suppose that $N$ is set using Algorithm \ref{algo1}, where $\varepsilon$ is given by 
\begin{equation}
\label{eps}
\varepsilon = \frac{w(0)  \delta}{2 T+\delta}.
\end{equation}
Then the output of Algorithm \ref{GPAcode} is within $\pm \delta$ of the output of the  bilateral filter. 
\end{corollary}

\section{Experiments and Discussion}
\label{exp}

\begin{figure}[b]
\centering
\subfloat[$I_1$]{\label{f1}\includegraphics[width=0.24\linewidth]{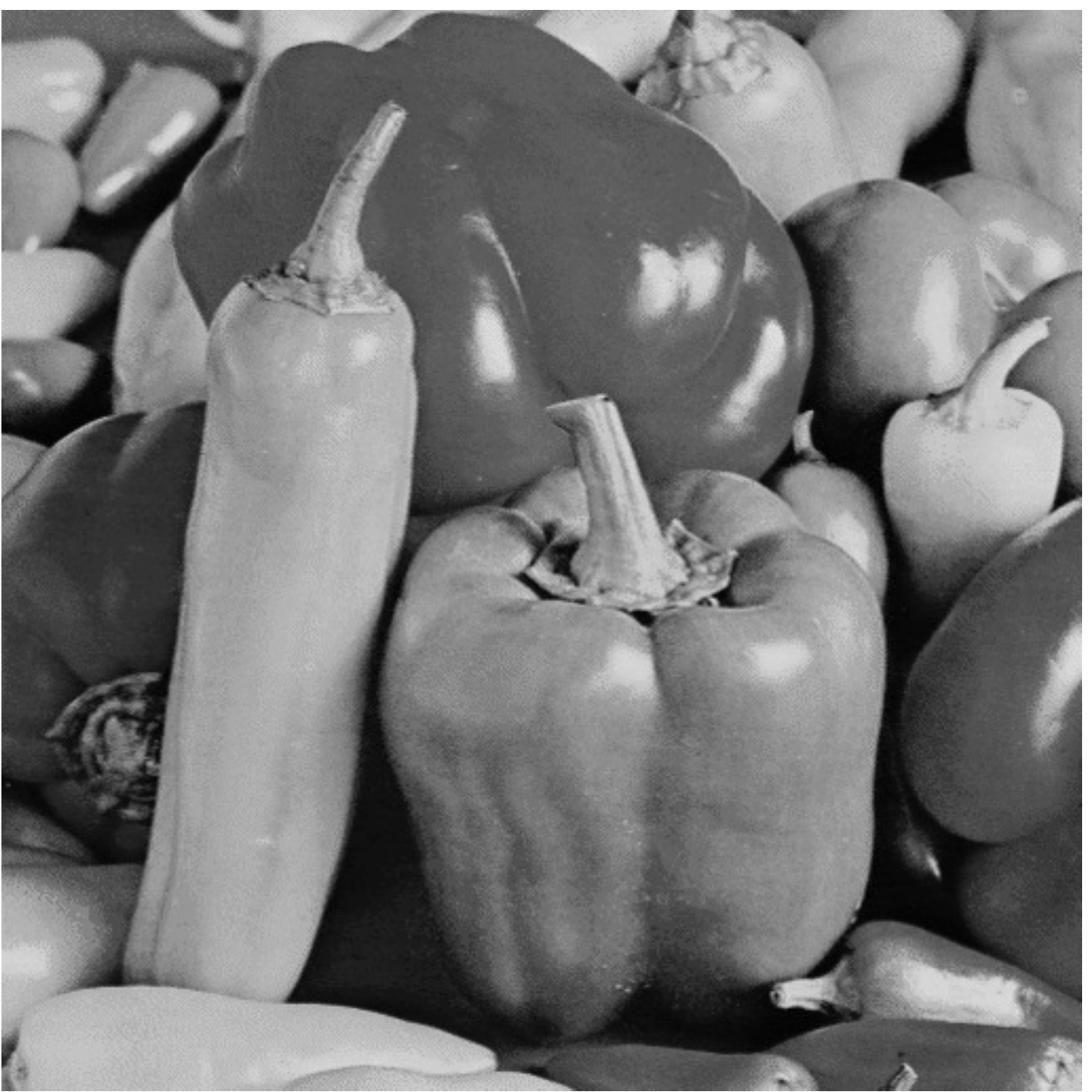}}  \
\subfloat[$I_2$]{\label{f2}\includegraphics[width=0.24\linewidth]{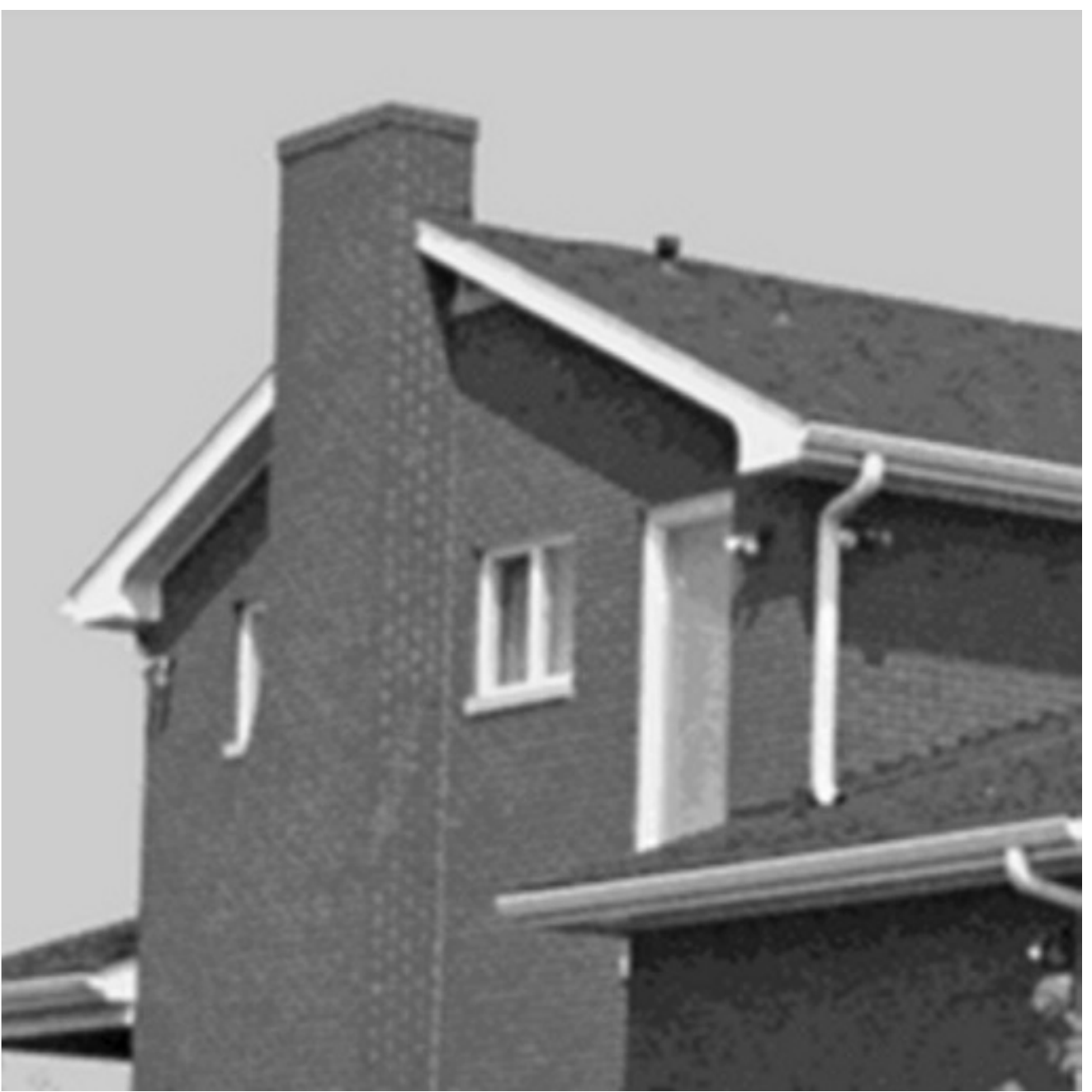}}  \
\subfloat[$I_3$] {\label{f3}\includegraphics[width=0.24\linewidth]{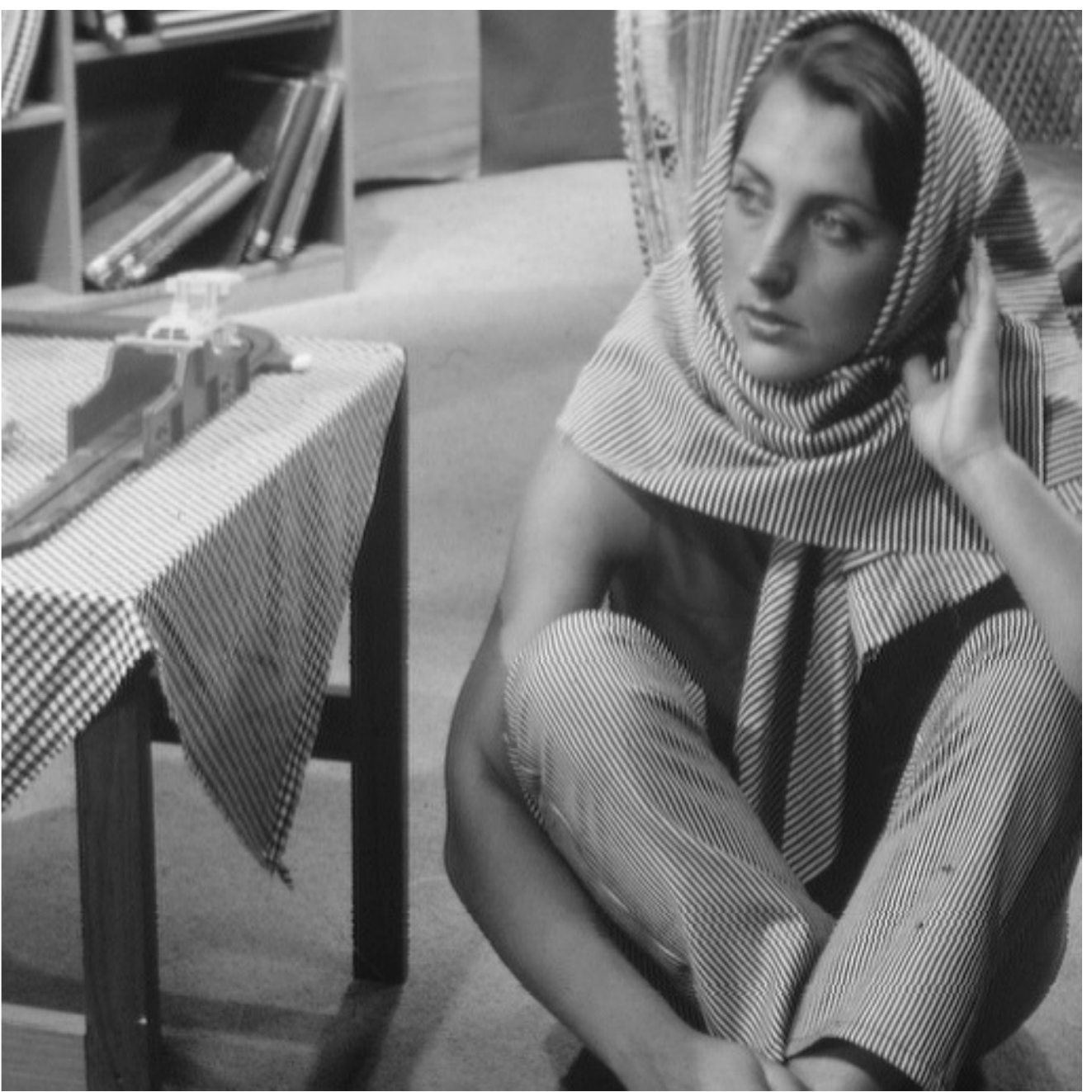}} \
\subfloat[$I_4$]{\label{f4}\includegraphics[width=0.24\linewidth]{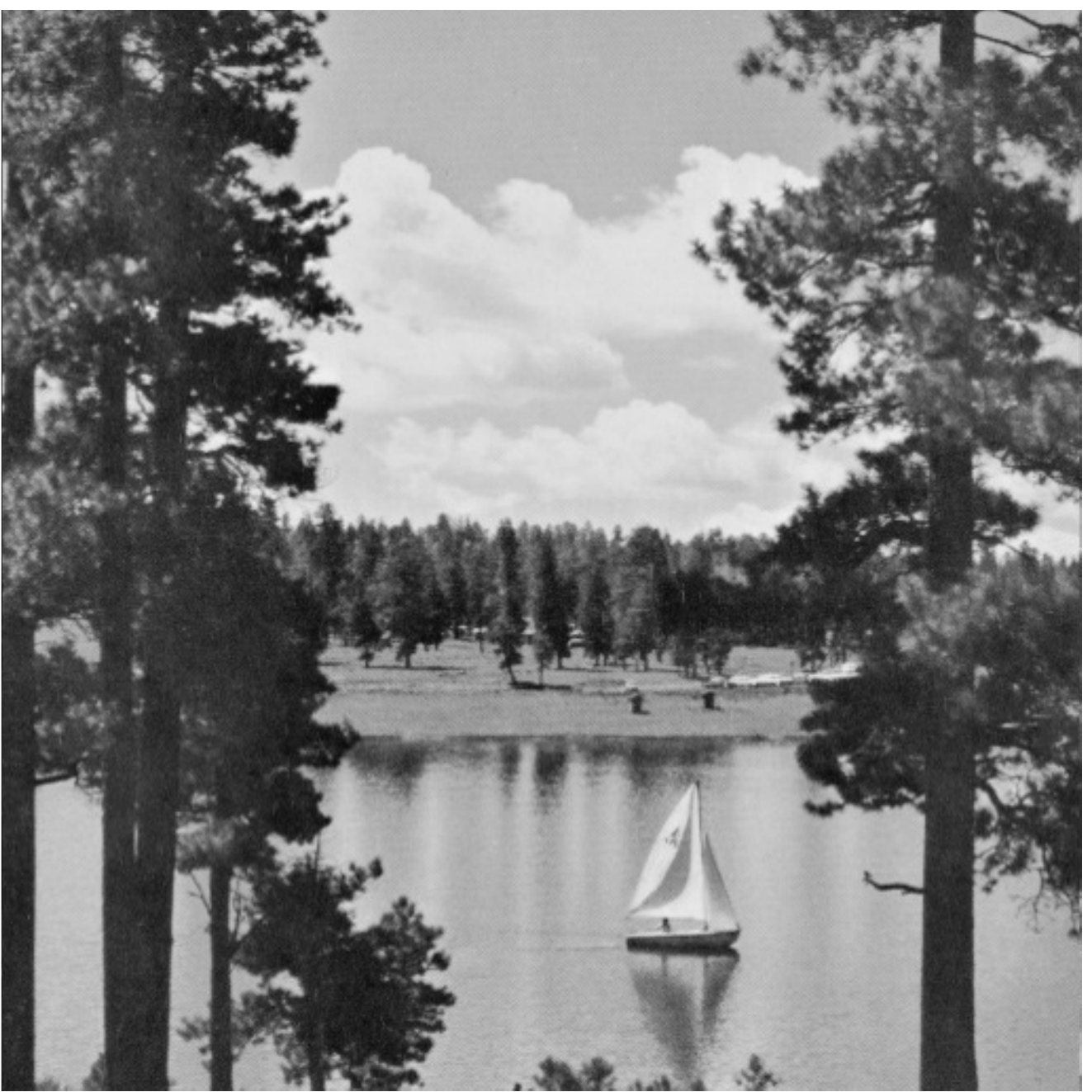}}  \\
\subfloat[$I_5$]{\label{f5}\includegraphics[width=0.24\linewidth]{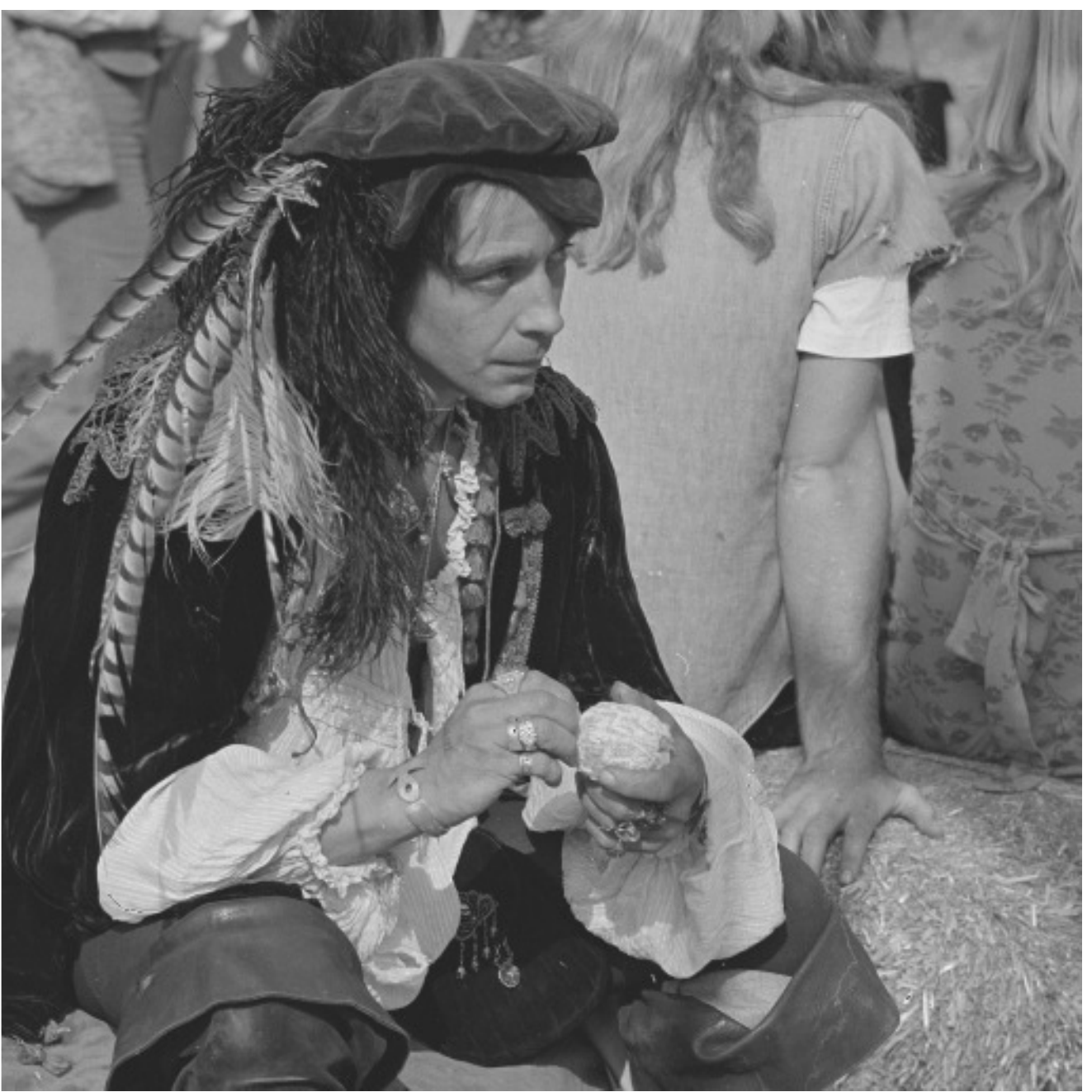}}  \
\subfloat[$I_6$]{\label{f6}\includegraphics[width=0.24\linewidth]{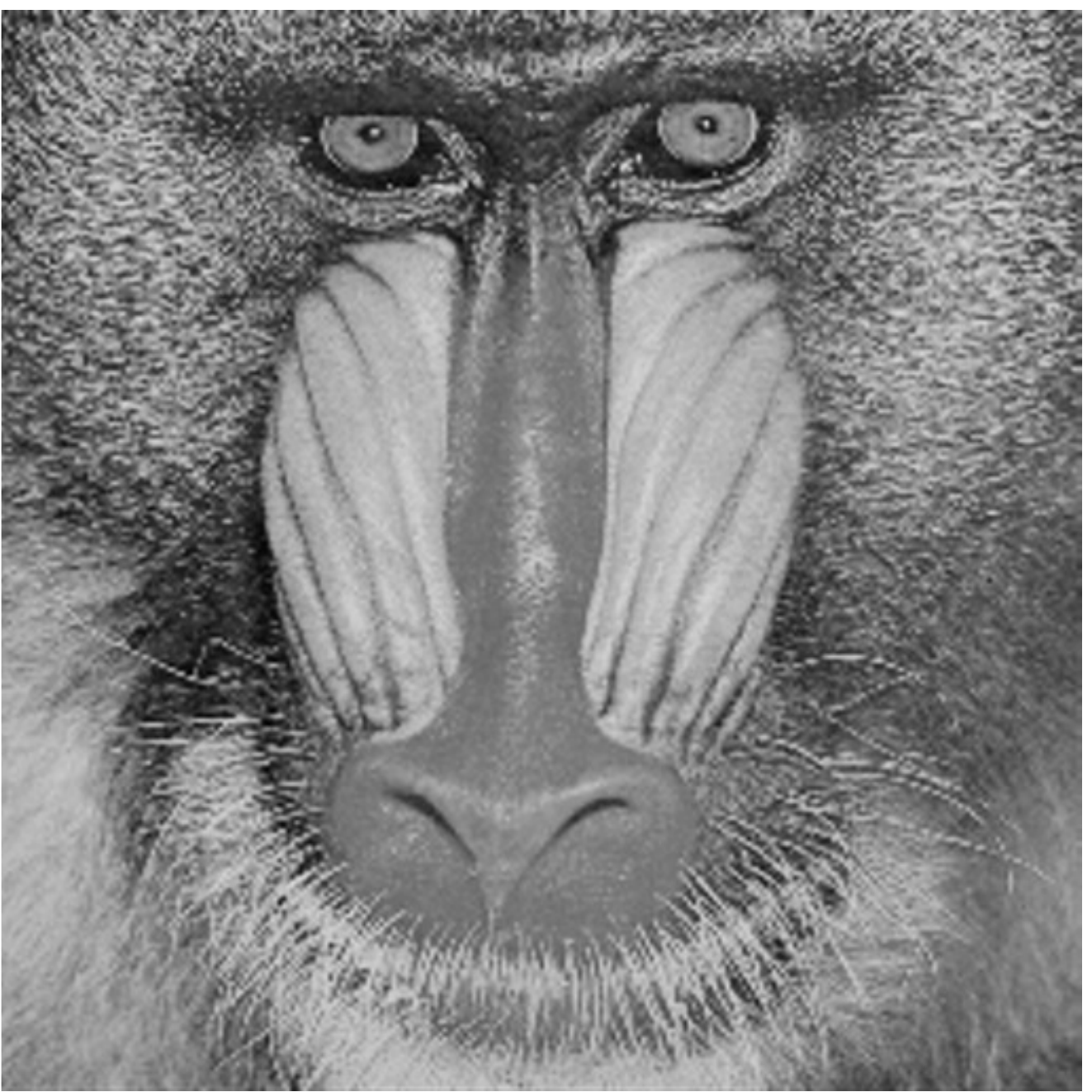}}  \
\subfloat[$I_7$]{\label{f7}\includegraphics[width=0.24\linewidth]{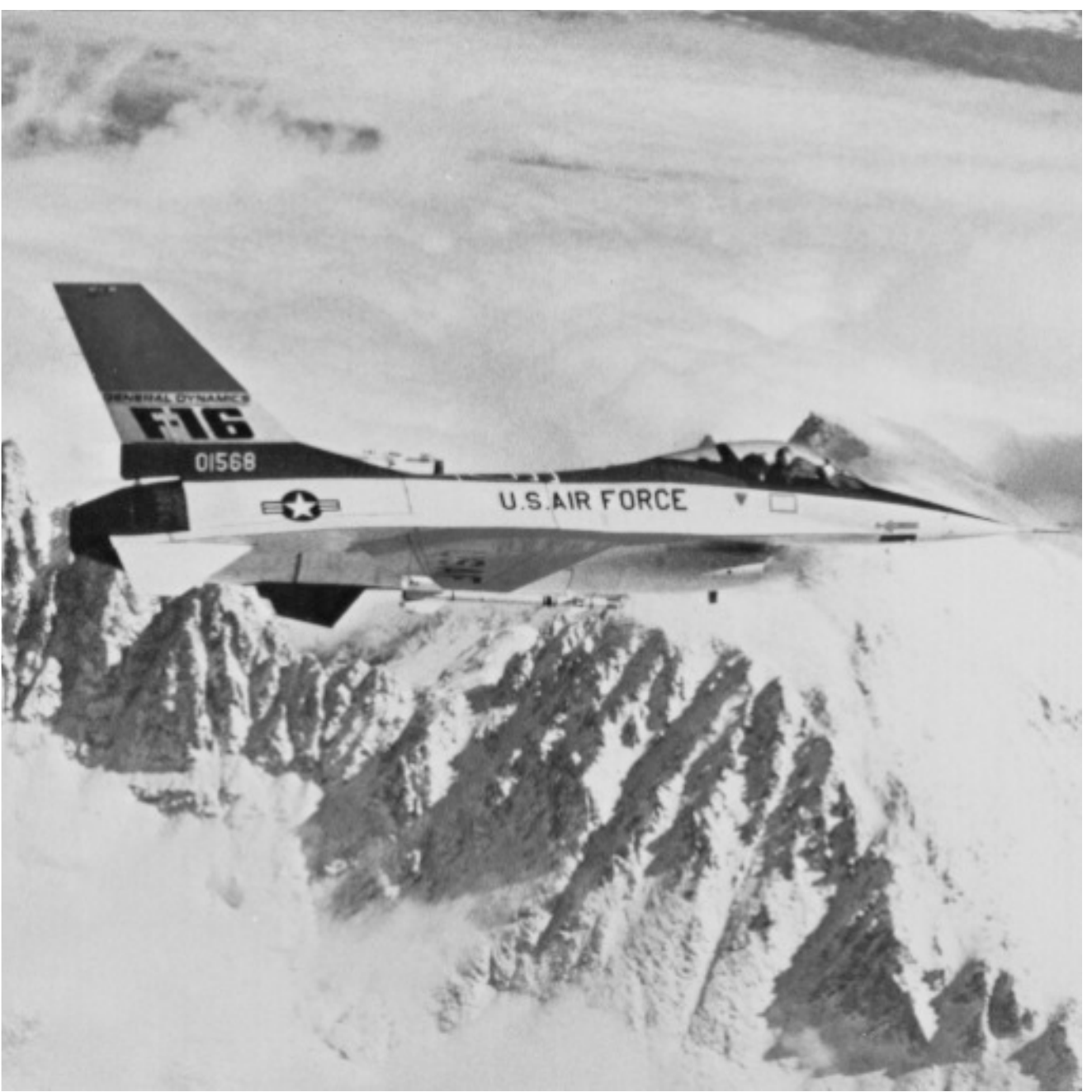}}  \
\subfloat[$I_8$]{\label{f8}\includegraphics[width=0.24\linewidth]{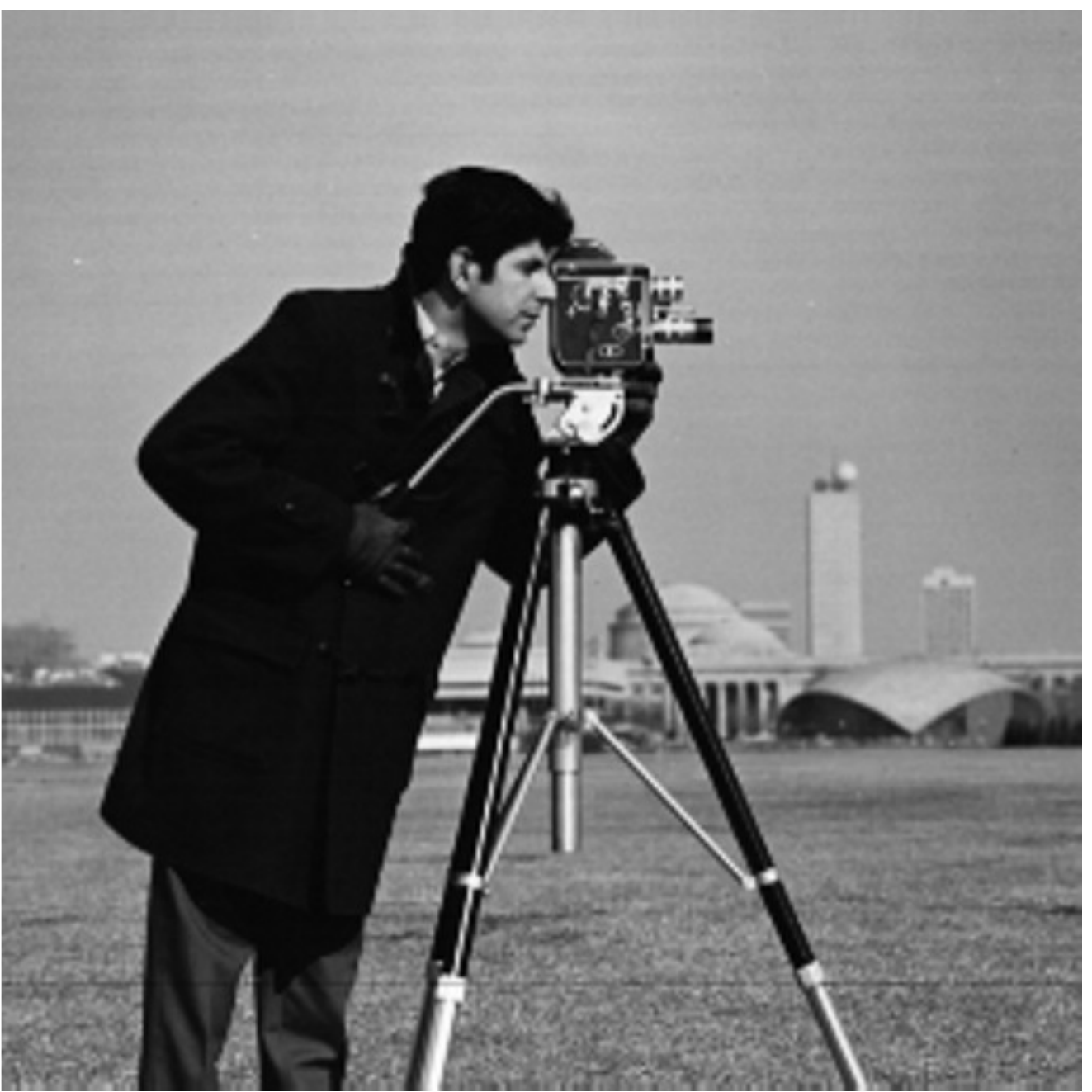}} 
\caption{List of grayscale images used for the experiments in Section \ref{exp}. The images were obtained from \cite{dataset}. All images are of size $512 \times 512$.}
\label{dataset}
\end{figure}

\begin{figure*}[t]
\subfloat[BF (10.2 sec).] {\includegraphics[width=0.24\linewidth]{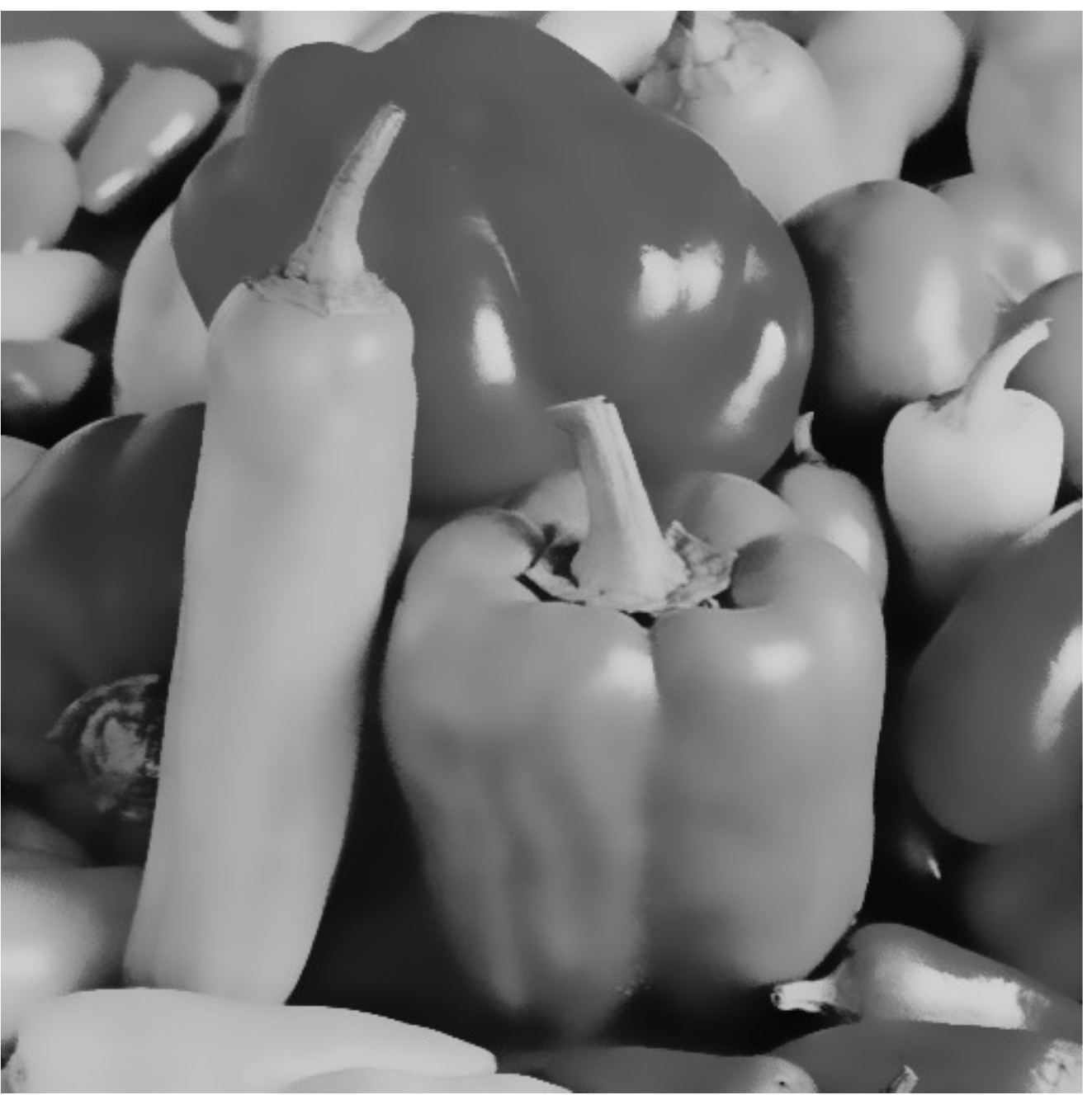}}  \
\subfloat[GPA (0.77 sec, -72 dB, -174 dB).] {\label{fig1}\includegraphics[width=0.24\linewidth]{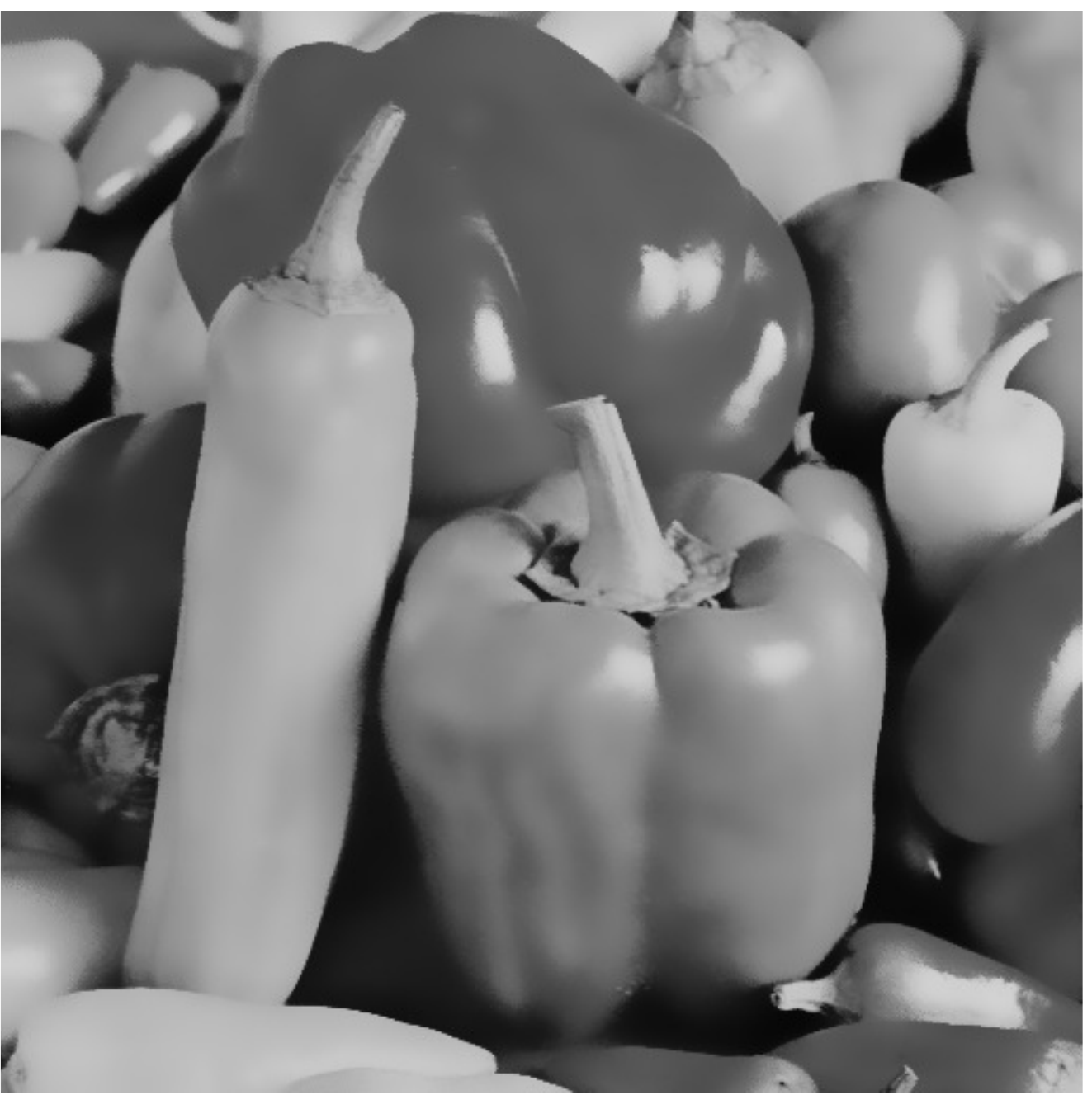}} \
\subfloat[BF (9.4 sec).] {\includegraphics[width=0.24\linewidth]{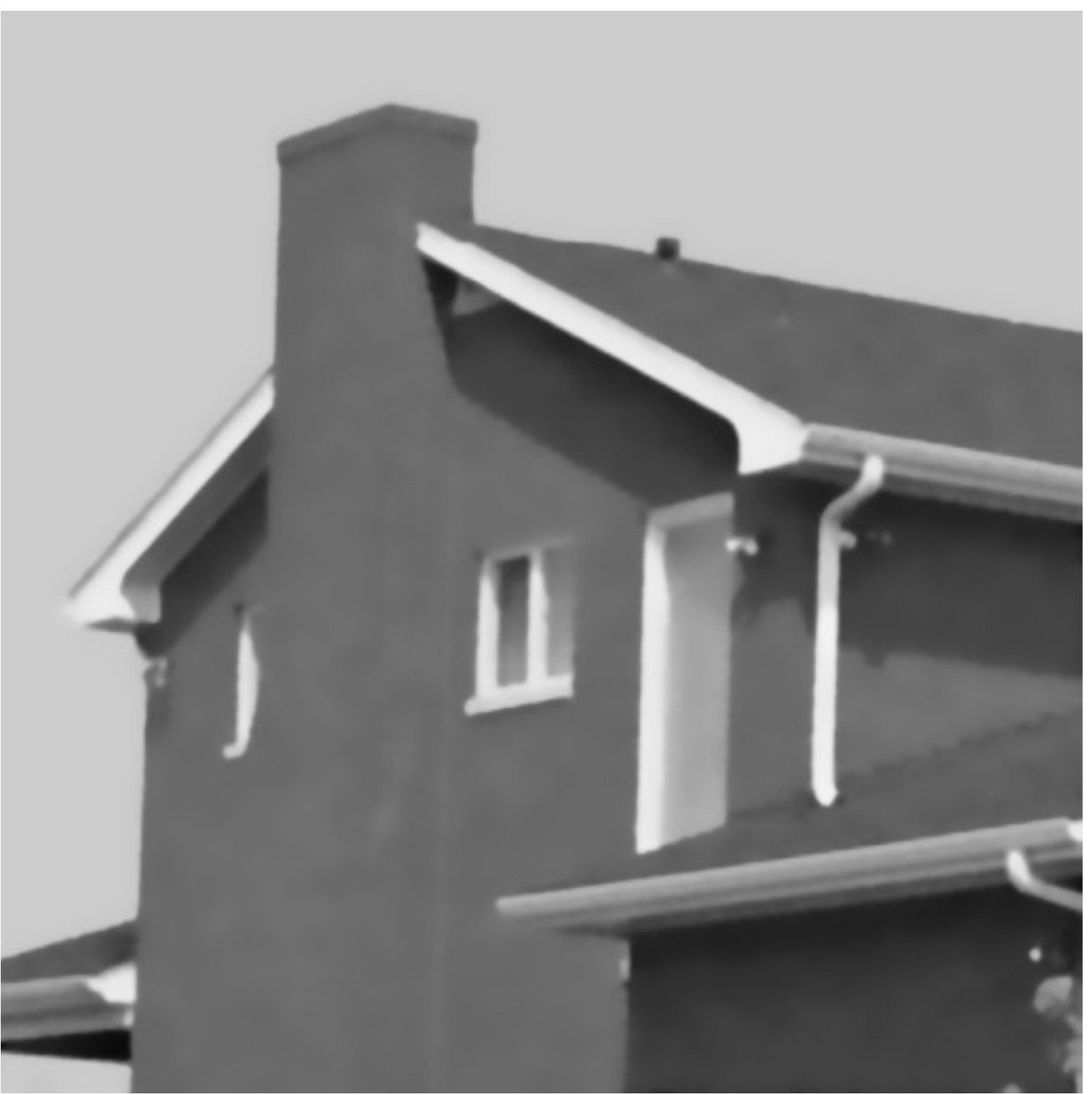}}  \
\subfloat[GPA (0.85 sec, -58 dB, -162 dB).] {\label{fig1}\includegraphics[width=0.24\linewidth]{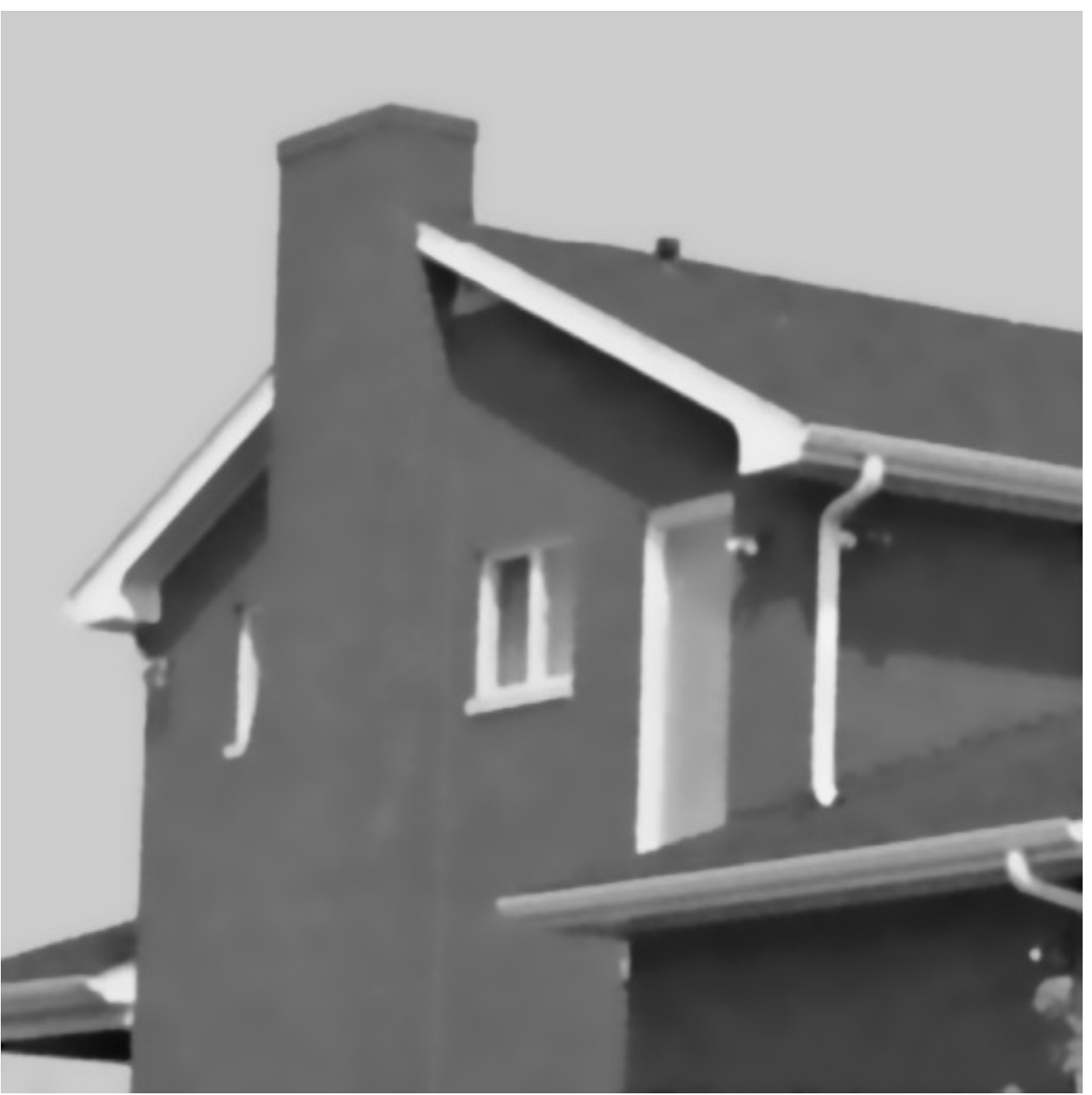}} 
 \caption{Comparison of the exact bilateral filter (BF) and the proposed approximation (GPA) on images $I_1$ and $I_2$. A Gaussian kernel ($\sigma_s=5$) is used for the spatial filter, and $\sigma_r=50$ for the Gaussian range kernel. The accuracy parameter was set to $\delta=0.1$ for the GPA. In the caption of (a) and (c), we report the run time of the BF. In the caption of (b) and (d), we report the run time of the GPA, and the $\ell_{\infty}$ and mean-squared errors between BF and GPA.}
\label{result1}
\end{figure*}

\begin{figure*}[t]
 \subfloat[BF (4.3 sec)] {\includegraphics[width=0.24\linewidth]{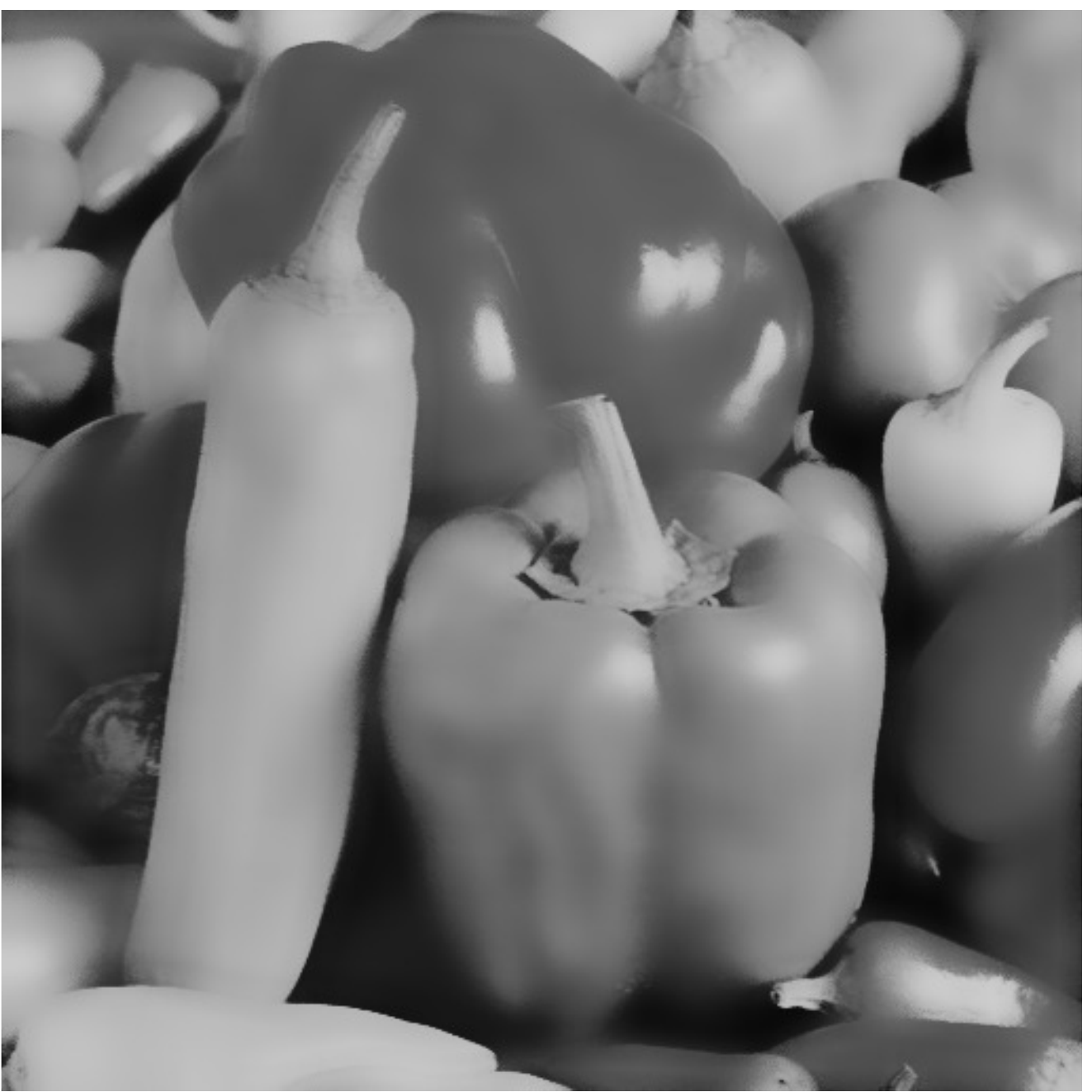}}  \
\subfloat[GPA (0.61 sec, -65 dB, -164 dB).] {\label{fig1}\includegraphics[width=0.24\linewidth]{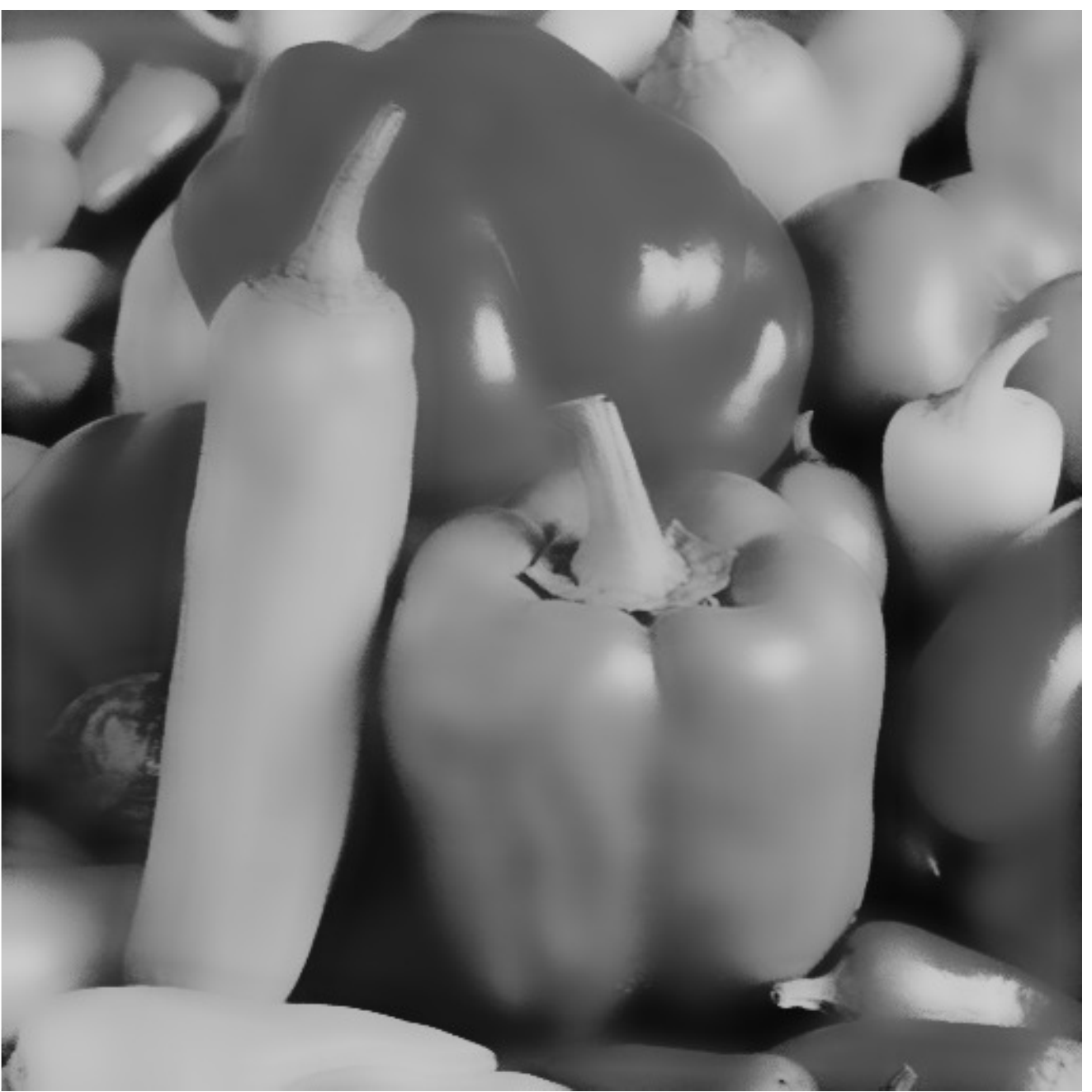}} \
\subfloat[BF (4.22 sec)] {\includegraphics[width=0.24\linewidth]{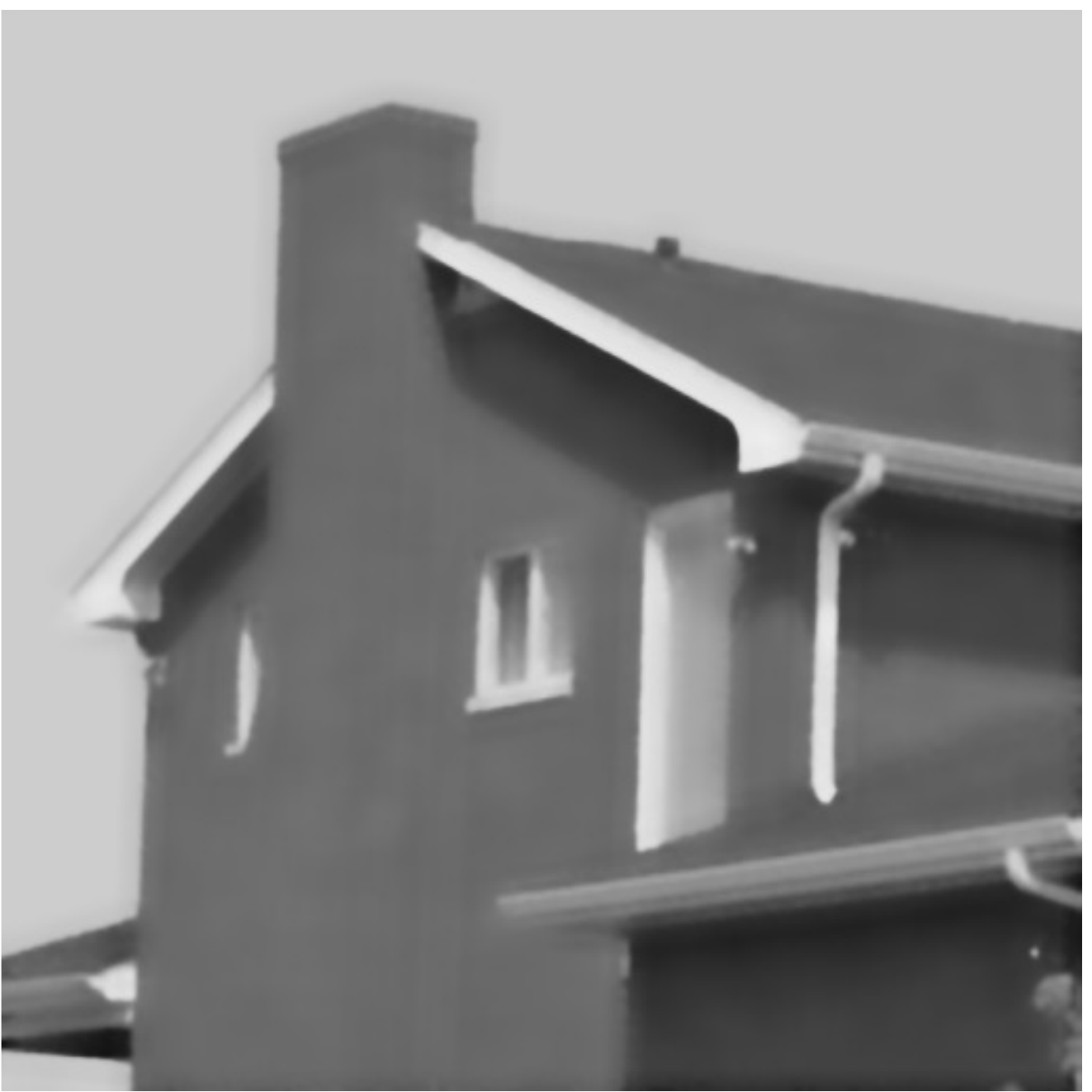}}  \
\subfloat[GPA (0.62 sec, -54 dB, -155 dB). ] {\label{fig1}\includegraphics[width=0.24\linewidth]{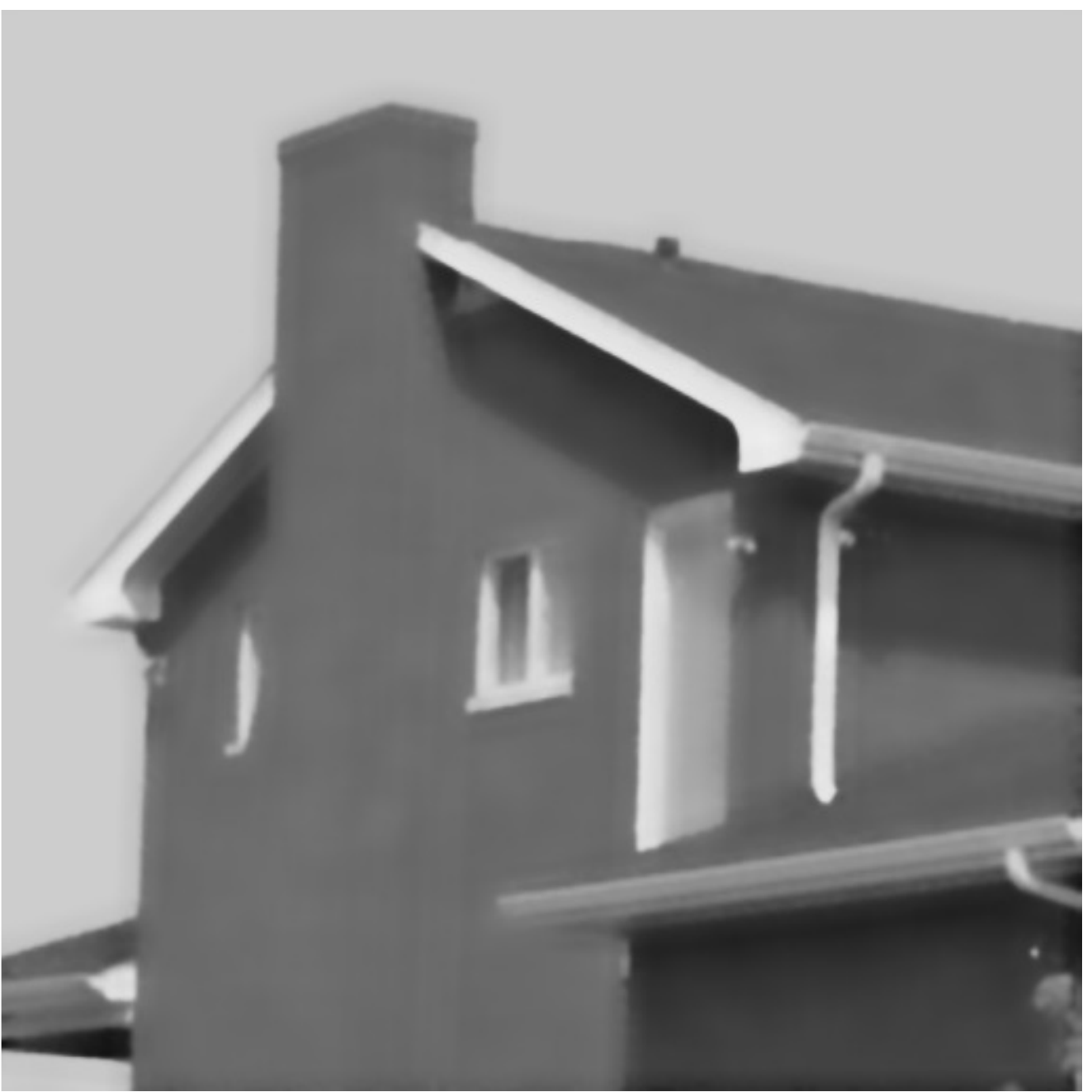}} 
 \caption{The setup here is identical to that in Figure \ref{result1}, with the difference that a box kernel  ($W = 10$) is used for the spatial filter instead of the Gaussian.}
\label{result2}
\end{figure*}

\begin{figure}[t]
\centering
 \subfloat[$\sigma_s=3$.]{\includegraphics[width=0.5\linewidth]{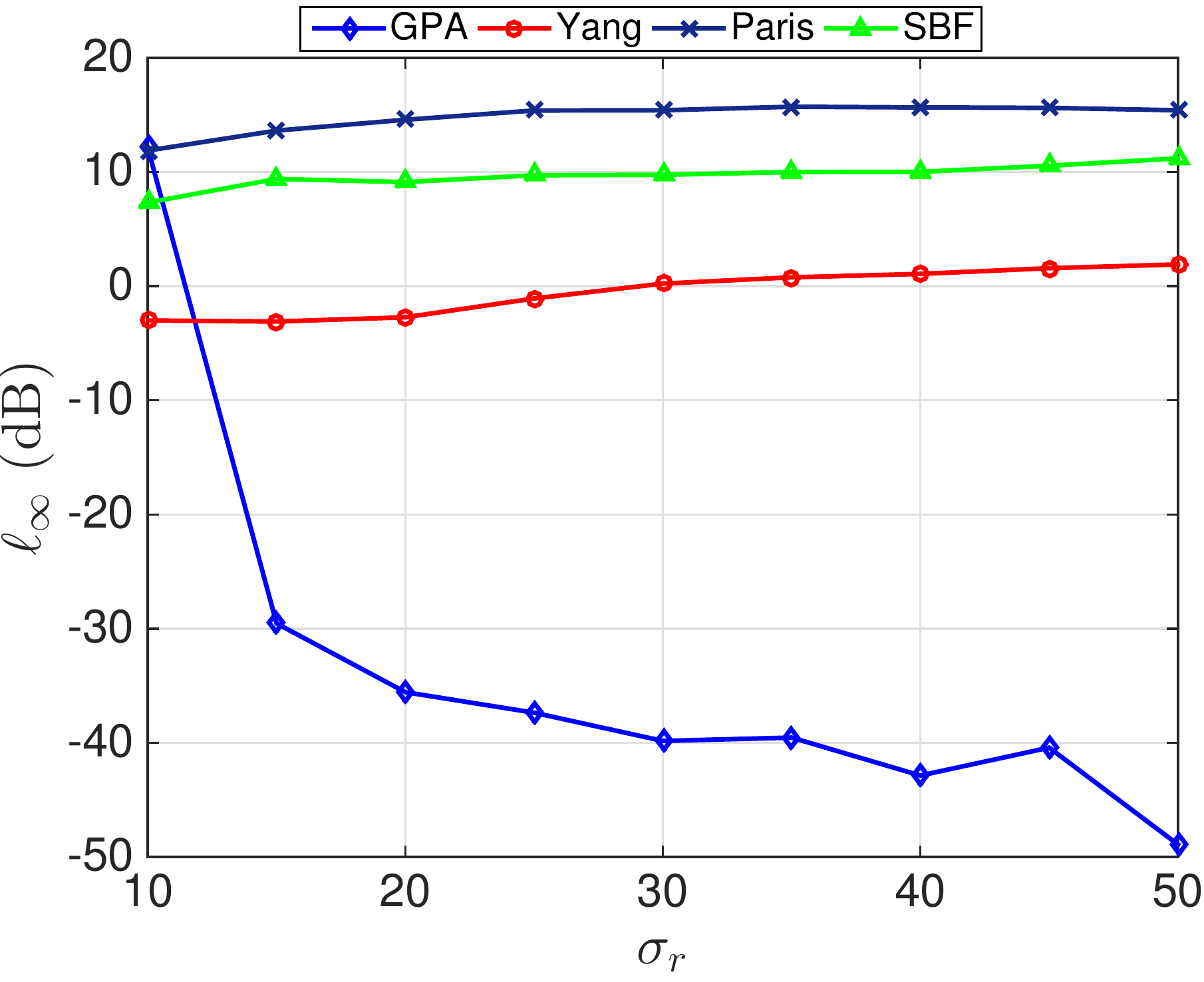}}  
 \subfloat[$\sigma_s=3$.] {\includegraphics[width=0.5\linewidth]{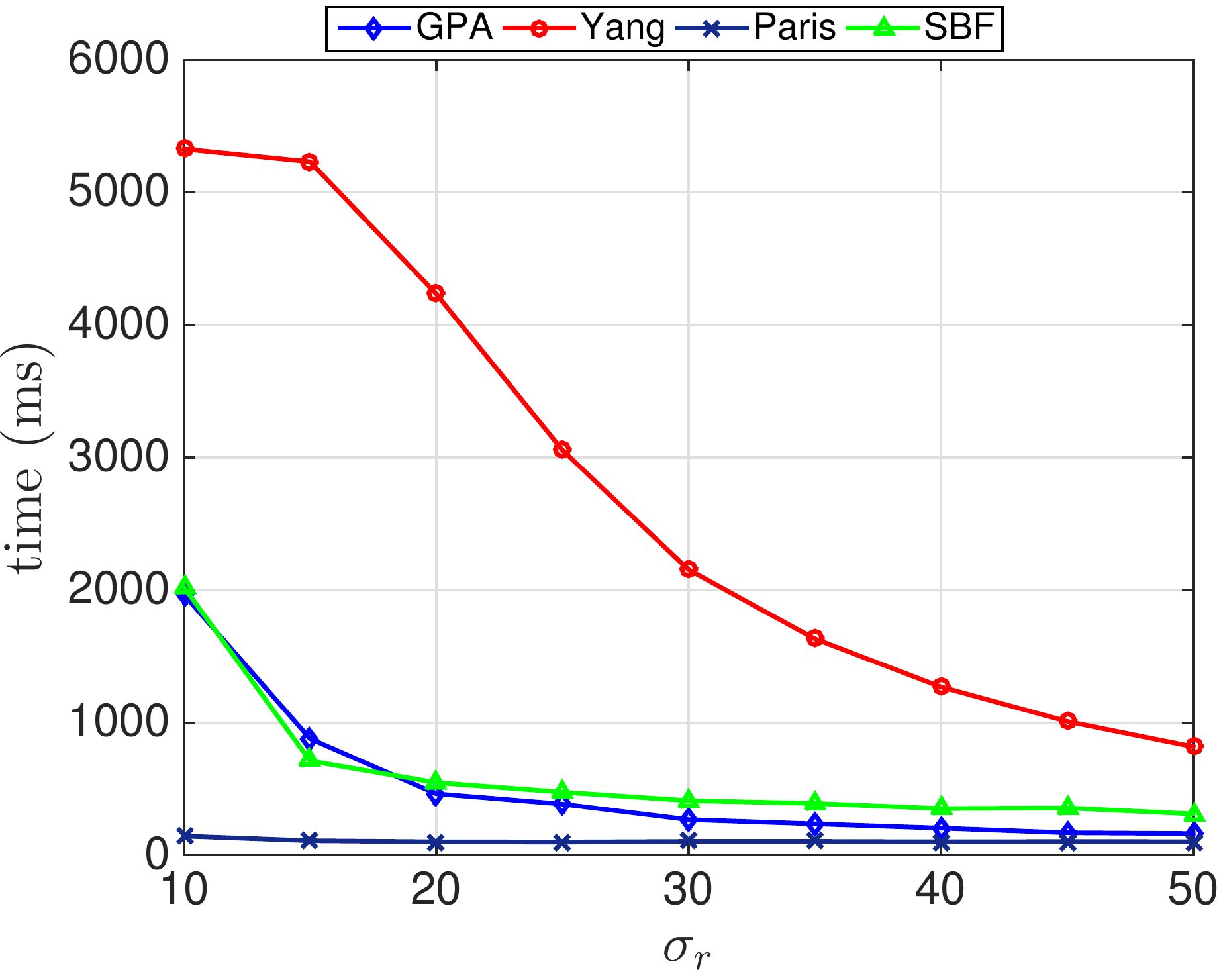}} \\
\subfloat[$\sigma_r=30$.]{\includegraphics[width=0.5\linewidth]{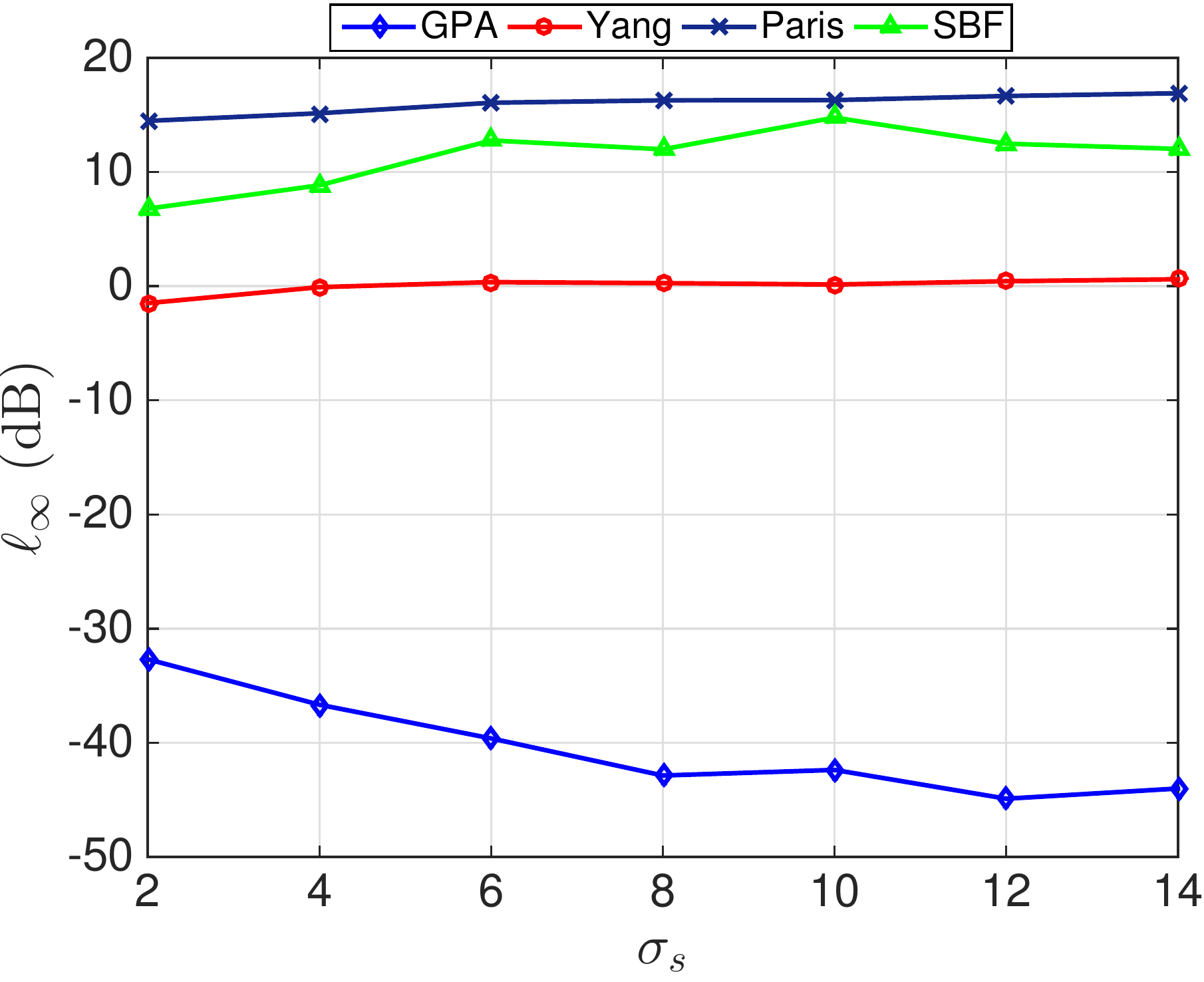}}  
\subfloat[$\sigma_r=30$.] {\includegraphics[width=0.5\linewidth]{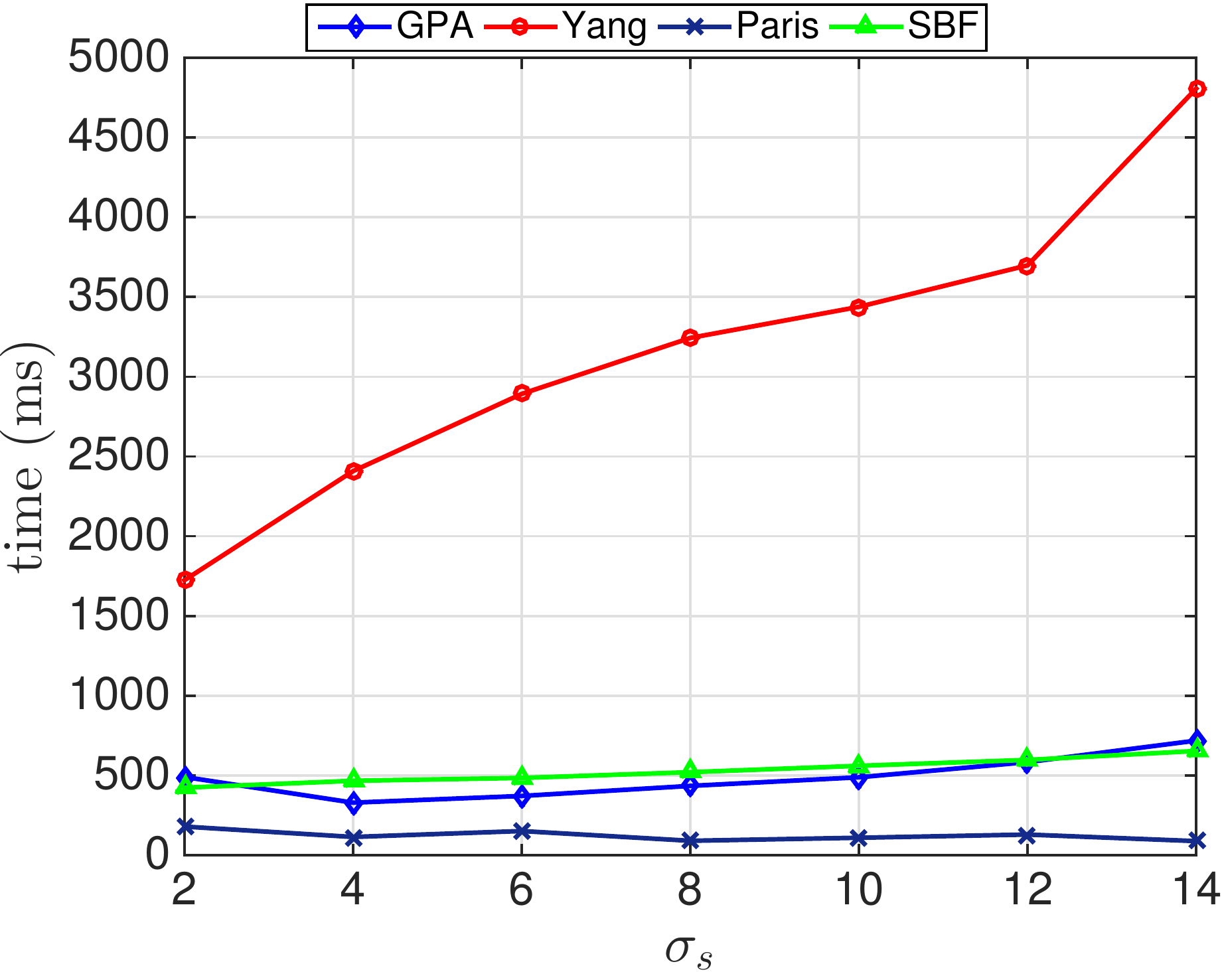}} 
\caption{Comparison of the filtering accuracy and the run time of four different algorithms as a function of the parameters $\sigma_s$ (Gaussian spatial filter) and $\sigma_r$. We used image $I_1$ in Figure \ref{dataset} for the comparison. We used $\delta=1$ for GPA and Yang's algorithm \cite{Yang2009}. A tolerance of $0.01$ was used for the SBF \cite{Chaudhury2013}.}
\label{result3}
\end{figure}

\begin{figure}[t]
\centering
\subfloat[$W=4$.]{\includegraphics[width=0.5\linewidth]{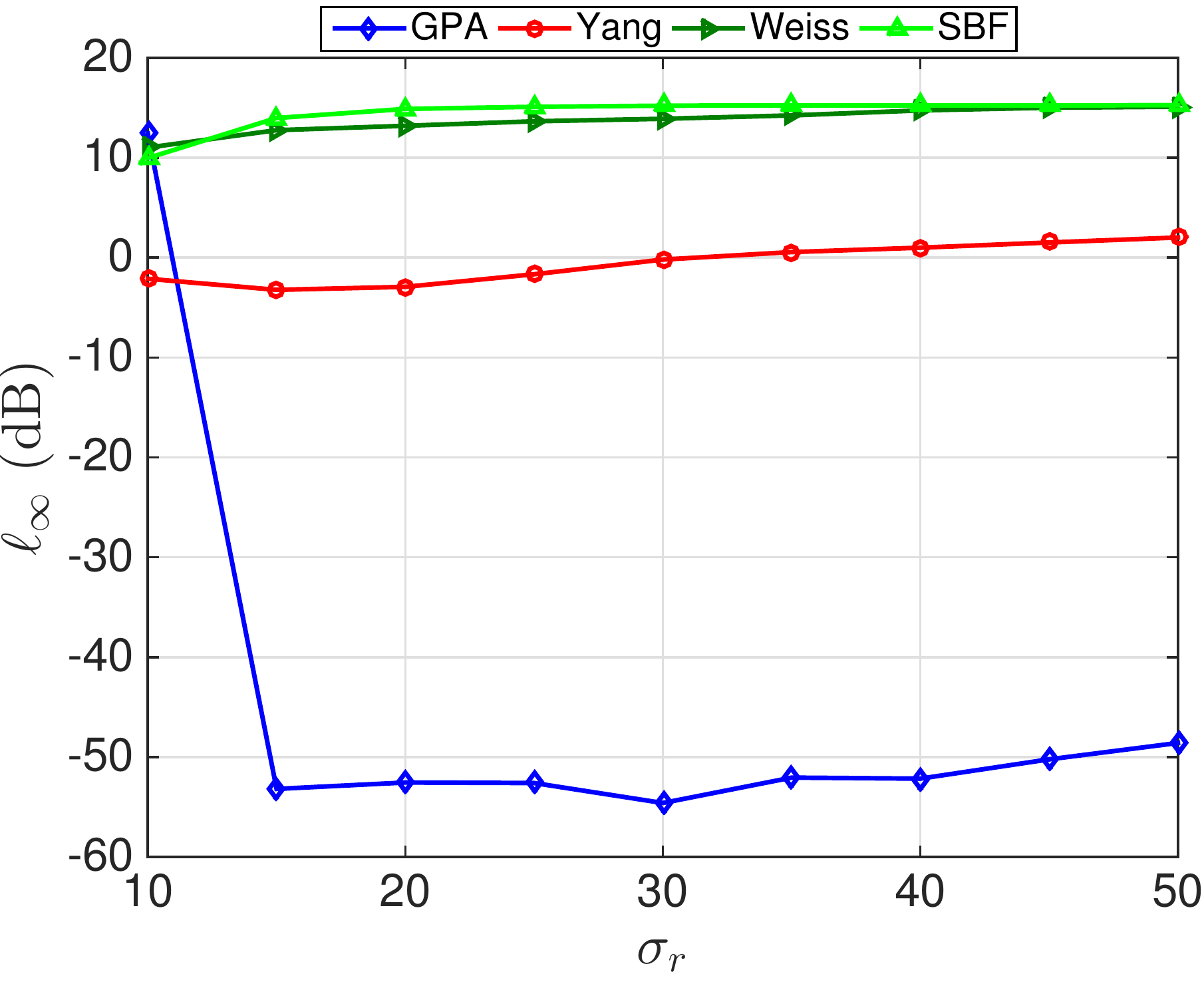}}  
\subfloat[$W=4$.]{\includegraphics[width=0.5\linewidth]{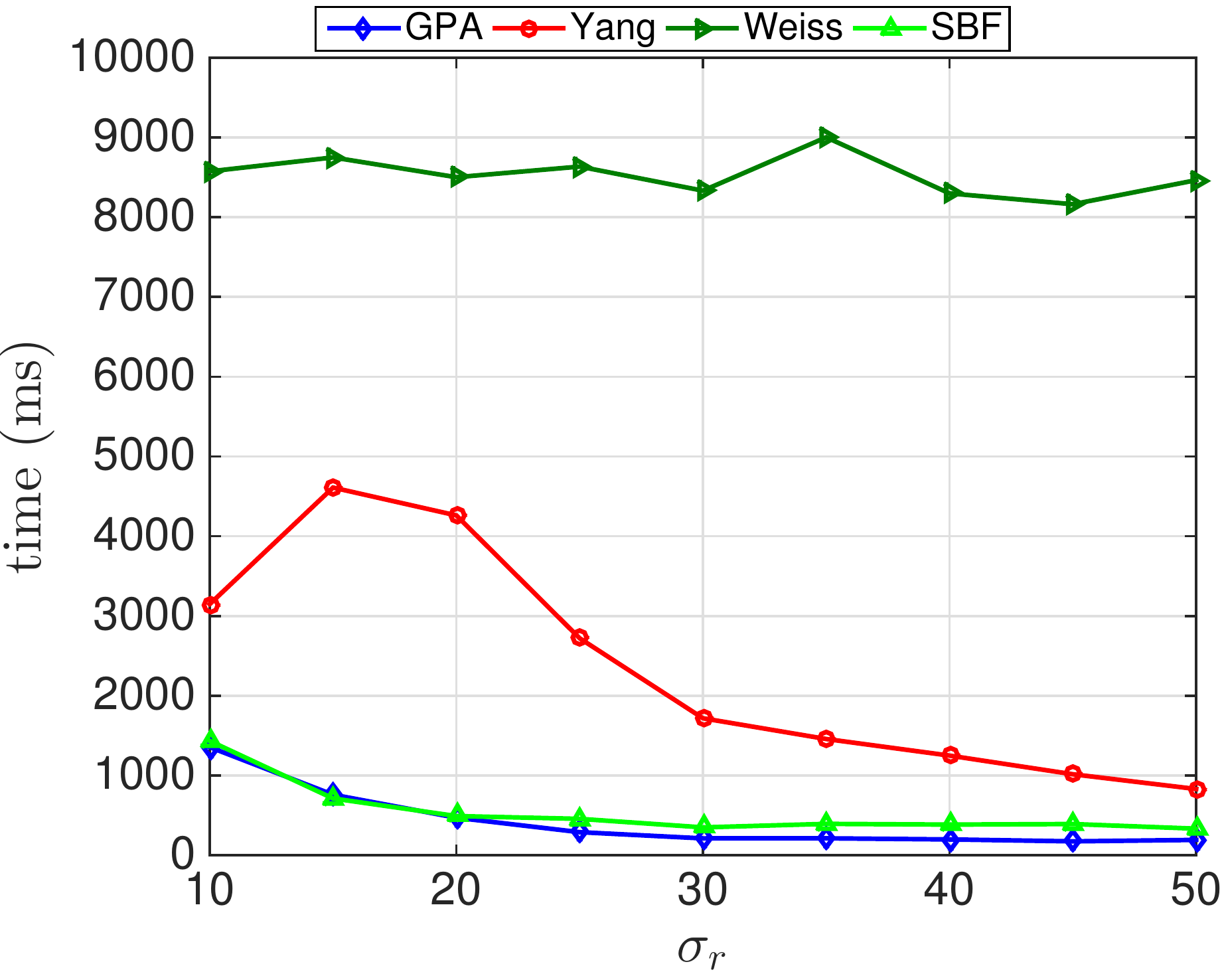}} \\
\subfloat[$\sigma_r=30$.]{\includegraphics[width=0.5\linewidth]{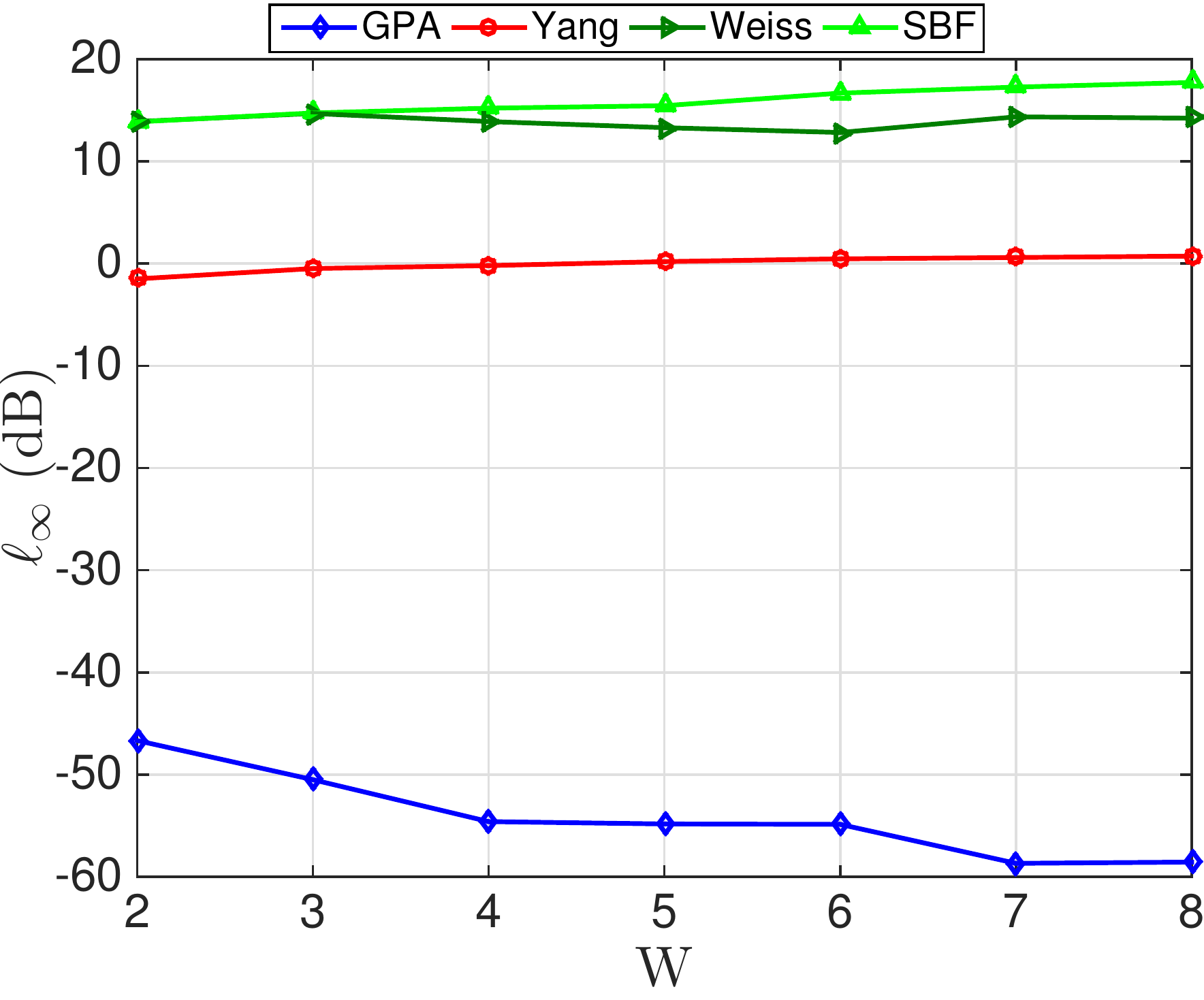}}  
\subfloat[$\sigma_r=30$.]{\includegraphics[width=0.5\linewidth]{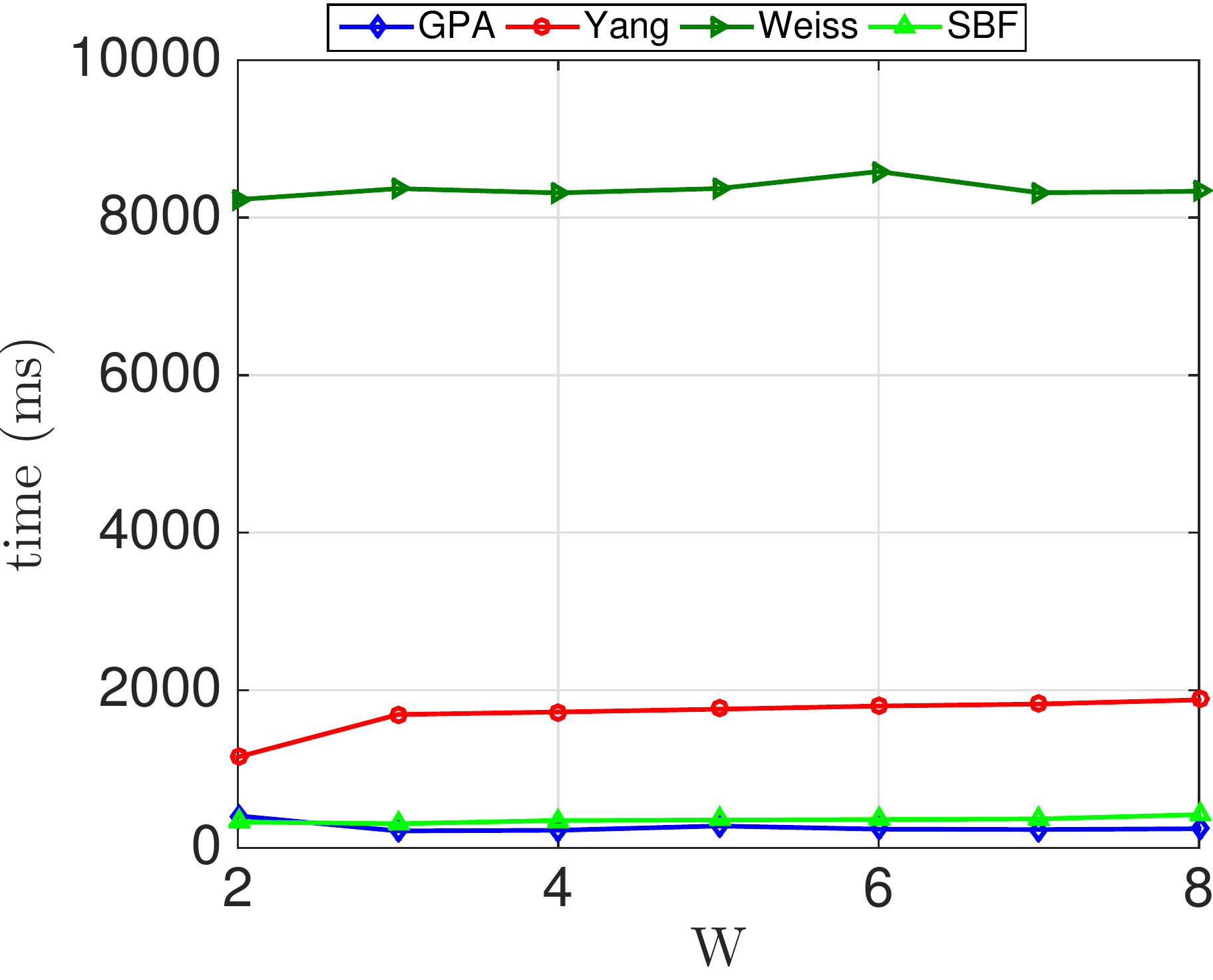}} 
\caption{Comparison of the filtering accuracy and the run time as a function of $W$ and $\sigma_r$. The settings are identical to that in Figure \ref{result3}; the difference here is that we have used a box spatial kernel instead of a Gaussian kernel.}
\label{result4}
\end{figure}

\begin{table}[b]
\centering
\caption{Comparison of the order $N_0$ required to achieve a desired accuracy $\delta$ when $\sigma_s=5$ and $\sigma_r=30$.}
\label{orderComp}
\begin{tabular}{|p{1.8cm} |p{0.6cm}|p{0.6cm}|p{0.6cm}|p{0.6cm}|p{0.6cm}|p{0.6cm}|p{0.6cm}|}
 \hline
$ \hspace{10mm} \delta$ &$10^{-3}$  & $10^{-2}$ & $0.05 $& $0.1$ &$1$ &$3$  \\ \hline
$[N_0]$ using \eqref{LambW}  &49  &46 &45 &44 &41 &40  \\ \hline
$[N_0]$ using \eqref{Nvsdelta} &49  &47 &45 &44 &42 &41  \\ \hline
$[N_0]$ using \eqref{Nest} &4006  &1267 &566 &401 &127 &73 \\
\hline
\end{tabular}
\end{table}

 \begin{figure*}
\centering
\subfloat[$\sigma_s=5$.]{\label{f1}\includegraphics[width=0.245\linewidth]{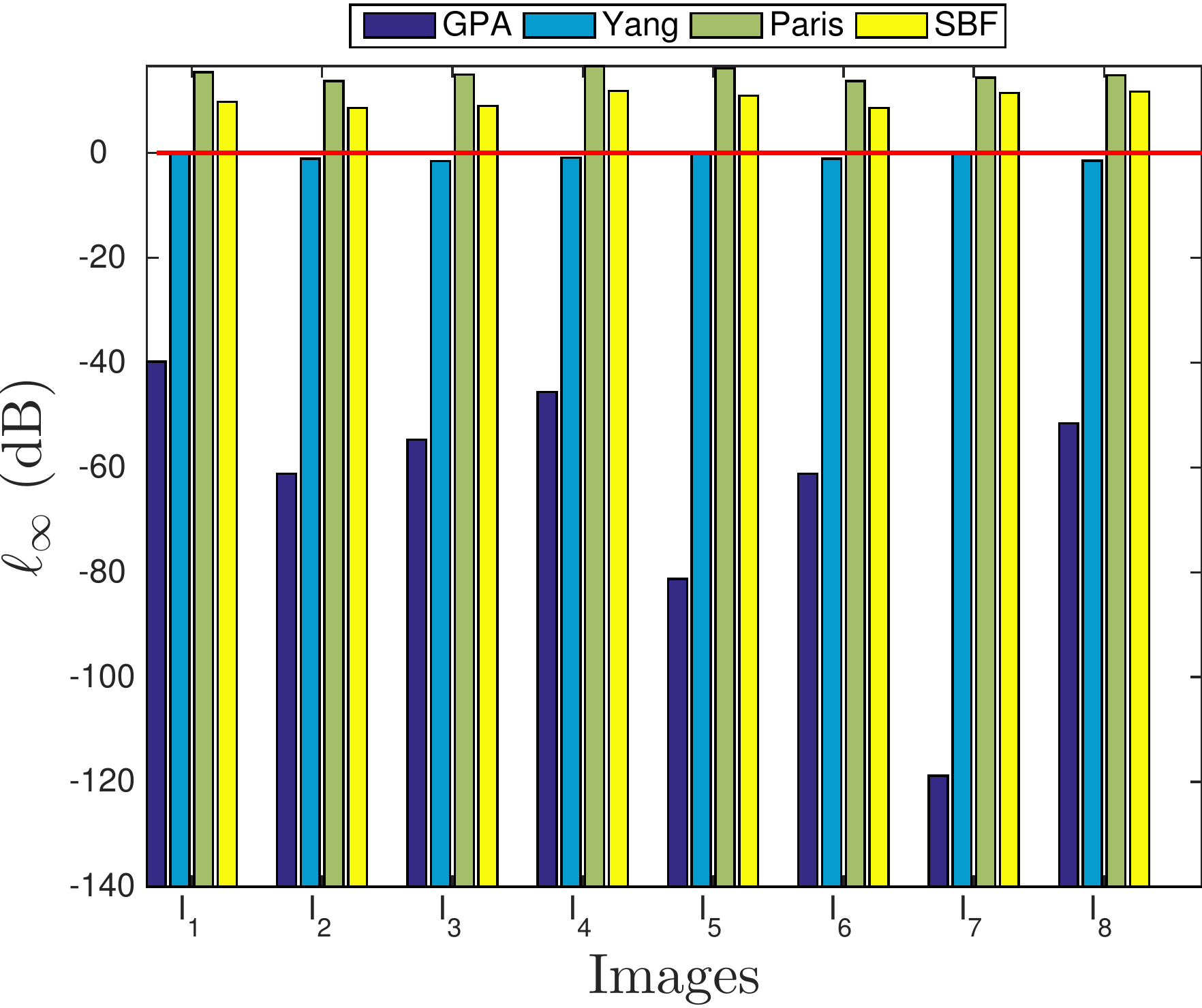}}  \
\subfloat[$\sigma_s=5$.]{\label{f2}\includegraphics[width=0.245\linewidth]{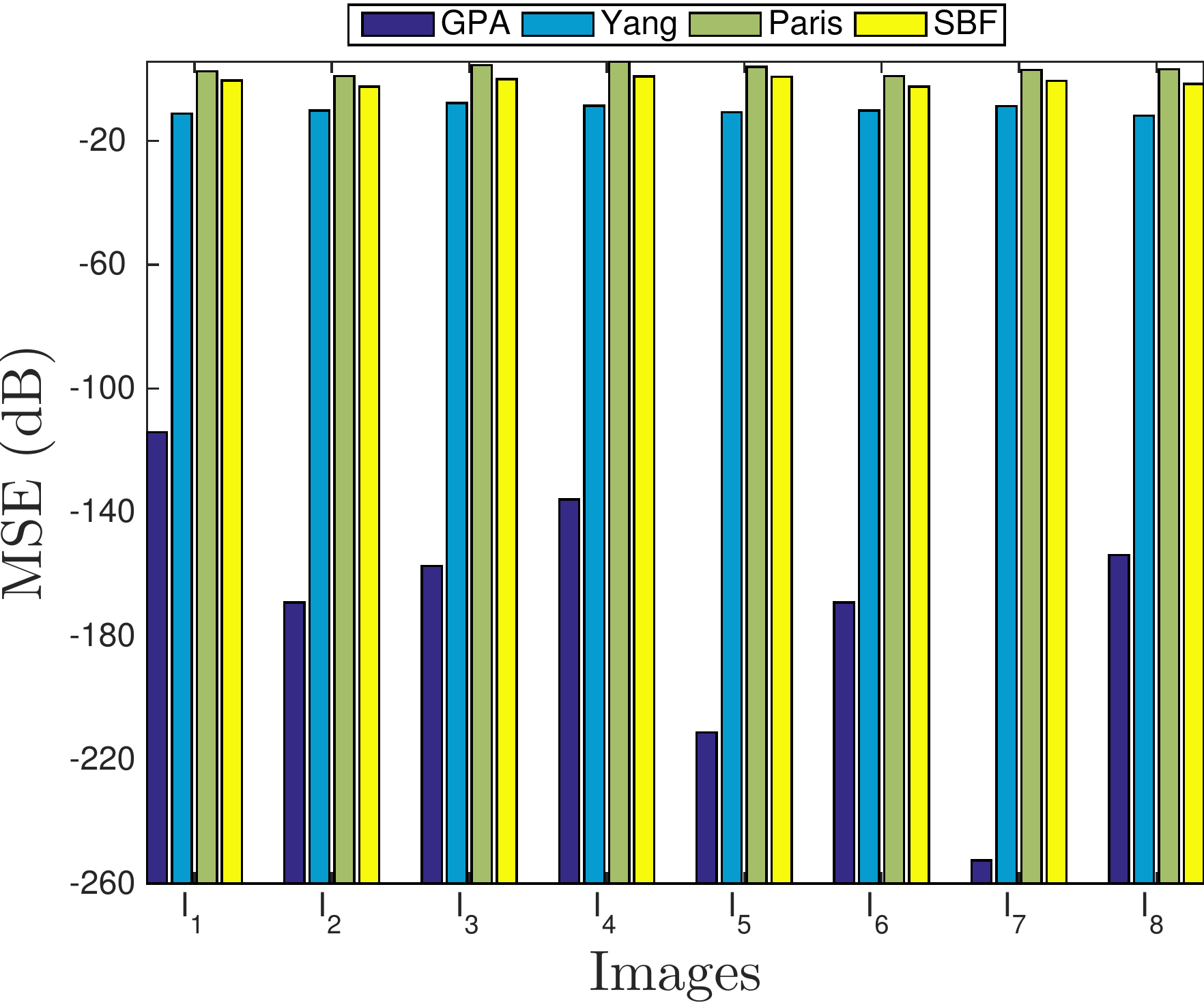}}  \
\subfloat[$W=4$.]{\label{f4}\includegraphics[width=0.245\linewidth]{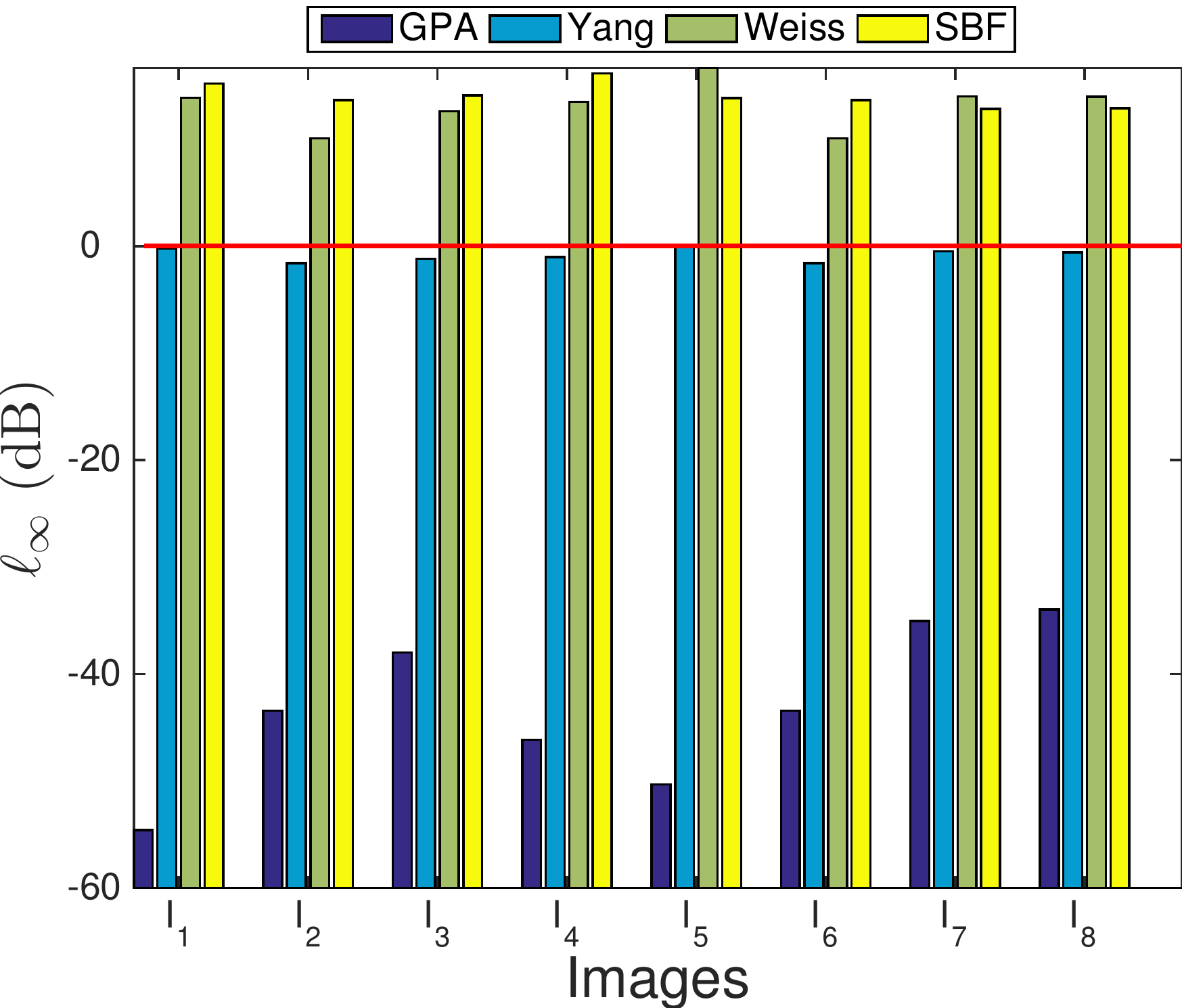}}  \
\subfloat[$W=4$.]{\label{f5}\includegraphics[width=0.245\linewidth]{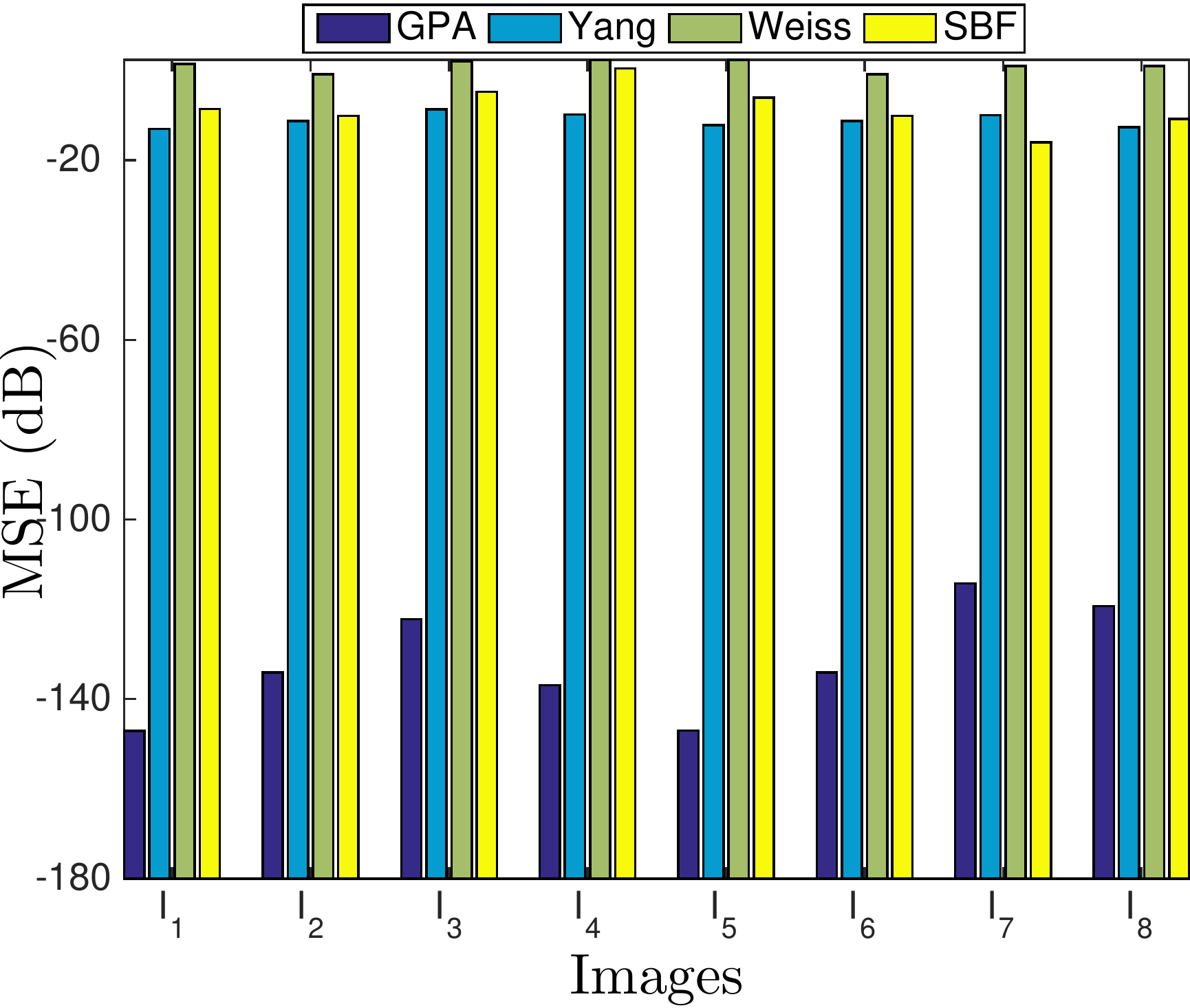}}  
\caption{Comparison of the filtering accuracy ($\ell_{\infty}$ error and MSE) of various fast algorithms on the images in Figure \ref{dataset}. The red horizontal lines in \ref{f1} and \ref{f4} represent the accuracy parameter $\delta$ used for GPA and Yang's algorithm. The tolerance for SBF was set to be $0.01$. In \ref{f1} - \ref{f2}, we show the results for a Gaussian spatial kernel, and in \ref{f4} - \ref{f5} we show the results for a box kernel. We used $\sigma_r=30$ for the Gaussian range kernel in all the experiments.}
\label{barPlot}
\end{figure*}

We implemented the proposed GPA algorithm using Matlab $8.4$ on an Intel $3.4$ GHz Linux system with $8$ GB memory. The Matlab implementation has been shared here \cite{Mcode}. The set of grayscale images used for the experiments are shown in Figure \ref{dataset}. We compared the proposed  algorithm with the following fast algorithms: Yang \cite{Yang2009}, Paris \cite{Paris2006}, Weiss \cite{Weiss2006}, and the Shiftable Bilateral Filter (SBF) \cite{Chaudhury2013}. We used the Matlab implementation of these algorithms to make the comparison fair;  moreover, we used the parameter settings suggested in the respective papers. For determining the order in \cite{Yang2009} for a given accuracy parameter $\delta$, we have used \eqref{Nest}.

\underline{\textbf{Experiment 1}} The output of the proposed GPA algorithm on a couple of  images are shown in Figures \ref{result1} and \ref{result2}.  We also provided the output obtained using exact bilateral filtering. We performed the comparison using the box and the Gaussian kernels for the spatial filter. Notice that the speedup obtained is significant. Moreover, the filtered images are visually identical and numerically very close, in terms of the  $\ell_{\infty}$ and mean-squared errors. We have used the following definition of mean-squared error (MSE):  
\begin{equation*}
\text{MSE}  = 10  \log_{10} \Big\{ |I|^{-1} \sum_{\i \in I} \big(f_{\mathrm{BF}}(\i) -  f_{\mathrm{GPA}}(\i) \big)^2 \Big\},
\end{equation*}
where $|I|$ denotes the number of pixels in the image.

To get a better understanding of how $N_0$ varies with $\delta$, we used the following approximation (see Appendix \ref{approxN0}): 
\begin{equation}
\label{Nvsdelta}
N_0 \approx  1.72 \left(\frac{T}{\sigma_r} \right)^2 \!+ \log\left(  \frac{2 T}{w(0)  \delta} \right).
\end{equation}
An important point to note in \eqref{Nvsdelta} is the logarithmic dependence on $\delta$. In fact, the $\log(1/\delta)$ factor can be traced back to the tail bound in \eqref{Chernoff}, which, in turn, follows from the particular splitting in \eqref{GaussPolynomial}. The implication of the logarithmic dependence is that we can force $\delta$ to be quite small without blowing up $N_0$.

To further highlight the importance of \eqref{Nvsdelta}, we compared \eqref{Nvsdelta} with the corresponding estimate  for Yang's algorithm \cite{Yang2009}:
\begin{equation}
\label{Nest}
N_0 \approx \frac{1.14 \times 10^5}{\delta^{1/2} \sigma_r^2}.
\end{equation}
The above estimate was recently derived in \cite{errbilat}. In particular, notice that the dependence on $\sigma_r$ is similar to that in \eqref{Nvsdelta}. However, the dependence on $\delta$ is much more strong in \eqref{Nest} compared to \eqref{Nvsdelta}, since $\log (1/\delta) \ll \delta^{-1/2}$ when $\delta < 1$.
 Moreover, the leading constant in \eqref{Nest} is much larger than the constant in the first term in \eqref{Nvsdelta}.
As an example, when $\delta=3$ and $\sigma_r=50$, we have $N_0 \approx 27$ for Yang's algorithm (this is the estimate reported in \cite{errbilat}). On the other hand, the corresponding estimate for our algorithm is $N_0 \approx 19$ (assuming that $T=128$ and that we use a box filter of size $3 \times 3$). The difference becomes dramatic for smaller values of $\delta$.
For example, when $\sigma_r=50$ and $\delta=0.01$, the estimate from \eqref{Nvsdelta} is $ 24$, while that from \eqref{Nest} is $456$. Further comparisons are provided in Table \ref{orderComp}. Notice that the order for Yang's approximation explodes when  $\delta < 1$ (sub-pixel accuracy). It is also seen from the table that \eqref{Nvsdelta} provides a close approximation of \eqref{LambW} for the setting under consideration.

\underline{\textbf{Experiment 2}} A graphic comparison of the algorithms for various settings of the spatial and range kernels is presented in Figures \ref{result3} and \ref{result4}. As before, we performed the comparison for both the box and Gaussian spatial filters. It is evident from these results that the proposed method is competitive with existing methods in terms of the speed-accuracy tradeoff.

\underline{\textbf{Experiment 3}} We next compared the proposed algorithm with existing fast algorithms on the images shown in Figure \ref{dataset}. A summary of the comparisons (in terms of maximum pixelwise error and MSE) is provided in Figure \ref{barPlot}.

\begin{table*}[t]
\centering
\caption{Comparison of the proposed GPA algorithm with Yang's algorithm \cite{Yang2009} for different order $N$. The $\ell_{\infty}$ error and the MSE are in decibels, while the time is in milliseconds. The comparison is done on image $I_5$ using both box and Gaussian spatial filters; the type of spatial filter is mentioned within brackets. The respective parameters for the box and Gaussian filter are $W=4$ and $\sigma_s=5$, and $\sigma_r=30$ for the Gaussian range kernel. Notice that the accuracy of GPA saturates above $N=60$.}
\label{table1}
\begin{tabular}{|l|l|l|l|l|l|l|l|l|l|l|l|l|}
\hline
$N$  & \multicolumn{3}{c|}{GPA (Gaussian)} & \multicolumn{3}{c|}{Yang (Gaussian)} & \multicolumn{3}{|c|}{GPA (Box)} & \multicolumn{3}{|c|}{Yang (Box)} \\
\hline
       & $\ell_{\infty}$ & MSE     & time                    & $\ell_{\infty}$ & MSE     & time & $\ell_{\infty}$ & MSE    & time & $\ell_{\infty}$ & MSE    & time \\
 \hline
  10    & 14.87     & 8.48  &  85   & 10.55     & 11.08     & 217 & 15.81     & 6.61  &  85   & 10.30     & 9.24  & 252 \\ \hline
  20    & 2.97  & -20.07    & 146   & 7.90  & 4.88  & 365 & 1.55  & -22.24    & 152   & 7.63  & 3.08  & 413 \\ \hline
  30    & -18.23    & -67.35    & 210   & 6.44  & 1.33  & 519 & -18.63    & -69.46    & 173   & 6.09  & -0.46     & 455 \\ \hline
  40    & -50.81    & -137.34   & 275   & 5.15  & -1.19     & 695 & -50.80    & -139.08   & 197   & 4.79  & -2.98     & 546 \\ \hline
  50    & -93.31    & -225.95   & 346   & 4.29  & -3.12     & 857 & -92.72    & -226.90   & 300   & 3.87  & -4.91     & 805 \\ \hline
  60    & -119.03   & -254.19   & 407   & 3.46  & -4.72     & 995 & -120.69   & -258.14   & 295   & 3.06  & -6.51     & 767 \\ \hline
  65    & -119.03   & -254.19   & 439   & 3.11  & -5.41     & 1067 &  -120.69   & -258.14   & 323   & 2.70  & -7.20     & 840 \\ \hline
  70    & -119.03   & -254.19   & 477   & 2.79  & -6.06     & 1175 & -120.69   & -258.14   & 456   & 2.38  & -7.85     & 1216 \\ \hline
\end{tabular}
\end{table*}

\begin{table*}
\centering
\caption{Comparison of the  GPA algorithm with Yang's algorithm \cite{Yang2009} at different $\delta$. See Table \ref{table1} for the parameter settings.}
\label{table2}
\begin{tabular}{|l|l|l|l|l|l|l|l|l|l|l|l|l|}
\hline
$\delta$ & \multicolumn{3}{|c|}{GPA (Gaussian)}                   &
\multicolumn{3}{c|}{Yang (Gaussian)}       & \multicolumn{3}{c|}{GPA
(Box)} & \multicolumn{3}{c|}{Yang (Box)}      \\
\hline
       & $N_0$ & $\ell_{\infty}$  & time       & $N_0$ & $\ell_{\infty}$
 & time & $N_0$ & $\ell_{\infty}$    & time & $N_0$  &
$\ell_{\infty}$   & time \\
       \hline
0.05 &    45 &  -71.15 &    445 &   567   & -6.29 &     11143 &         
44 & -66.51 &    429 &  567  &  -6.71 &     8905   \\ \hline
0.1 &    44 &   -66.89 &    383 &   401   & -4.78 &     8407 &      43 &
-62.45 &    207 &   401  &  -5.20 &     5381 \\ \hline
0.5 &     42 &  -58.65 &    376 &   180   & -1.29 &     3776 &    41 &
-54.59 &    282 &  180   & -1.71 &     3092\\ \hline
1 &  41 & -54.68 &    370 &  132   & -0.22 &  2635 &     41 & -54.59 &   
288 &   132 & -0.20 &     1981 \\ \hline
2 &  41 & -54.68 &    377 &   90  & 1.71 &  1692 &     40 & -50.80 &   
192 &   90 &   1.30 &  1137 \\ \hline
3 &  40 & -50.81 &    295 &   74  & 2.55 &  1305 &    39 & -47.10 &    262
&   74  &  2.14 &  1246 \\ \hline
\end{tabular}
\end{table*}

\underline{\textbf{Experiment 4}} Finally, we performed a detailed comparison of the proposed algorithm with Yang's algorithm, which is widely considered to be the state-of-the-art algorithm. In the first comparison, we fixed an image and the parameters of the bilateral filter. The order $N$ was then varied and the corresponding error and run times were noted. The results are presented in Table \ref{table1}. Notice that the run time of GPA is consistently smaller than that of Yang's algorithm for both the box and Gaussian kernels. Indeed, as remarked earlier, for a fixed order $N$, Yang's algorithm \cite{Yang2009} requires  $2N$ spatial filterings, while GPA requires only $N+1$ spatial filterings. Thus, the runtime of GPA is about half of that of Yang's algorithm. Moreover, beyond a certain $N$, GPA provides much better filtering accuracy. We performed a similar experiment by varying $\delta$, the results of which are reported in Table \ref{table2}. Notice that the run time of Yang's algorithm becomes prohibitively large when $\delta$ is small.

\section{Conclusion}
\label{conc}

We presented a novel fast algorithm for approximating the bilateral filter. The algorithm was shown to be both fast and accurate in practice using extensive experiments. 
The space and time complexity of the proposed algorithm is smaller than the state-of-the-art algorithm of Yang \cite{Yang2009}, and, moreover, was shown to provide much better accuracy.
We also performed an error analysis of the approximation scheme, and presented a rule for setting the approximation order that can guarantee the filtering accuracy to be within a desired margin. 
Before concluding, we note that the proposed algorithm can be used to perform cross bilateral filtering, and can also be extended for the filtering of video and volume data.

\section{Appendix}

\subsection{Derivation of \eqref{LambW}}
\label{proof1}

Taking the logarithm of \eqref{inequality}, we can restate the problem as one of finding the smallest integer $x > \lambda$ such that
\begin{equation}
\label{root}
\nu(x)=x \log x - px - q \geq 0
\end{equation}
where $p=1+\log(\lambda)$ and $q=-\lambda-\log \varepsilon$. 

Notice that $\nu'(\lambda)=0$ and $\nu''(x)=1/x >0$. Hence, $\nu(x)$ is strictly convex over $(0,\infty)$ with a minimum at $x=\lambda$. Since $\nu(\lambda) = \log \varepsilon < 0$ when $\varepsilon <1$, we conclude that there exists some $\theta > \lambda$ for which $\nu(\theta) = 0$. The smallest integer solution of \eqref{inequality} is precisely $[\theta]$. To find $\theta$, we solve the equations $\nu(\theta)=0$ and $\theta > \lambda$. Note that we can write $\nu(\theta)=0$ as
\begin{equation}
\frac{q}{\theta}\exp\left(\frac{q}{\theta} \right)=q e^{-p},
\end{equation}
which is of the form $y \exp(y) = q e^{-p}$, where $y=q/\theta$. The inverse of the mapping $y \mapsto y \exp(y) $ is a well-studied function called the Lambert W-function \cite{lamb}. In particular, the inverse (which is generally multivalued) in this case is given by
\begin{equation*}
\frac{q}{\theta}= W_0 (q e^{-p}),
\end{equation*}
where $W_0(t)$ is one of the two branches of the Lambert W-function \cite{lamb}. This gives us estimate \eqref{LambW}.

\subsection{Derivation of \eqref{PbyQ}}
\label{fastAlgo}

In terms of \eqref{intImg}, we can write $\phi_{N,\sigma_r}(f(\i-\j)-f(\i)) f(\i-\j)$ as
\begin{equation}
\label{expansion}
  \sigma_r  \exp\left(- \frac{f(\i)^2}{2\sigma_r^2}\right) \sum_{n=0}^{N-1} \frac{1}{n!} G_n(\i) F_{n+1}(\i - \j).
\end{equation}
On substituting \eqref{expansion} in the numerator of \eqref{GPA}, and exchanging the summations, we get
\begin{equation*}
\sum_{\j \in \Omega} \! w(\j)   \phi_{N,\sigma_r} \!(f(\i-\j)-f(\i)) f(\i-\j) = \exp\left(\!\! - \frac{f(\i)^2}{2\sigma_r^2}\right) \! P(\i),
\end{equation*}
which gives us \eqref{P} where we have used \eqref{spatialFiltering}. Similarly, on substituting \eqref{expansion} in the denominator of \eqref{GPA}, and exchanging the summations, we get 
\begin{equation*}
\sum_{\j \in \Omega} w(\j)  \phi_{N,\sigma_r}(f(\i-\j)-f(\i)) =\exp\left(- \frac{f(\i)^2}{2\sigma_r^2}\right)Q(\i),
\end{equation*}
where $Q(\i)$ is given by \eqref{Q}. Cancelling the common exponential term from the numerator and denominator, we get \eqref{PbyQ}.

\subsection{Derivation of \eqref{accuracy}}
\label{proof2}

To establish \eqref{accuracy}, we write \eqref{BF} as $f_{\mathrm{BF}}(\i)=  P_1(\i)/Q_1(\i)$, where 
\begin{equation*}
P_1(\i)=\sum_{\j \in \Omega} w(\j) \  g_{\sigma_r}(f(\i-\j)-f(\i)) \ f(\i-\j),
\end{equation*}
and
\begin{equation*}
Q_1(\i)= \sum_{\j \in \Omega} w(\j)  \  g_{\sigma_r}(f(\i-\j)-f(\i)).
\end{equation*}
Similarly, we write \eqref{GPA} as $f_{\mathrm{GPA}}(\i)=  P_2(\i)/Q_2(\i)$, where 
\begin{equation*}
P_2(\i)=\sum_{\j \in \Omega} w(\j) \  \phi_{N,\sigma_r}(f(\i-\j)-f(\i)) \ f(\i-\j),
\end{equation*}
and
\begin{equation*}
Q_2(\i)= \sum_{\j \in \Omega} w(\j)  \  \phi_{N,\sigma_r}(f(\i-\j)-f(\i)).
\end{equation*}
We can then write $f_{\mathrm{BF}}(\i) -  f_{\mathrm{GPA}}(\i)$ as
\begin{align}
\label{split}
&= \frac{P_1(\i) (Q_2(\i)-Q_1(\i)) + Q_1(\i) (P_1(\i)-P_2(\i))}{Q_1(\i)Q_2(\i)} \nonumber \\
 & = \frac{1}{Q_2(\i)} \Big[f_{\mathrm{BF}}(\i) (Q_2(\i)-Q_1(\i)) + P_1(\i)-P_2(\i) \Big].
\end{align}
We uniformly upper-bound (resp. lower-bound) the numerator (resp. denominator) in \eqref{split}. In particular, note that 
\begin{equation}
\label{UB}
\lVert  f_{\mathrm{BF}}  \rVert_{\infty}  \leq T.
\end{equation}
This follows from the fact that $f_{\mathrm{BF}} (\i)$ in \eqref{BF} can be expressed as a convex combination of $\{f(\i- \j) : \j \in \Omega\}$. On the other hand, $Q_2(\i)-Q_1(\i)$ is
\begin{equation*}
\sum_{\j \in \Omega} w(\j)  \left[ g_{\sigma_r}(f(\i-\j)-f(\i)) - \phi_{N,\sigma_r}(f(\i-\j)-f(\i)) \right].
\end{equation*}
Therefore, using \eqref{normal}, we get
\begin{equation}
\label{Qgap}
\lVert Q_1 - Q_2 \rVert_{\infty}\leq \lVert  E_{N,\sigma_r} \rVert_{\infty}.
\end{equation}
Similarly,
\begin{equation}
\label{Pgap}
\lVert P_1 -P_2 \rVert_{\infty}\leq \lVert  E_{N,\sigma_r} \rVert_{\infty}  T.
\end{equation}
To uniformly lower-bound $Q_2(\i)$, we note that for $\i \in I$,
\begin{equation*}
Q_1(\i) = w(0) g_{\sigma_r}(0)  + \!\! \! \sum_{\j \in \Omega \backslash \{0\}} \!\! \! w(\j)   g_{\sigma_r}(f(\i-\j)-f(\i)) \geq w(0),
\end{equation*}
where we have used the non-negativity of the range and spatial kernels. Using the inverse triangle inequality along with \eqref{Qgap}, we have for $\i \in I$,
\begin{equation}
\label{lb}
|Q_2(\i)| \geq Q_1(\i) -  |Q_2(\i) - Q_1(\i)| \geq   w(0) - \lVert  E_{N,\sigma_r} \rVert_{\infty}.
\end{equation}
Combining \eqref{split} - \eqref{lb}, we arrive at \eqref{accuracy}.

\subsection{Derivation of \eqref{Nvsdelta}}
\label{approxN0}

Note that typically $\delta \ll  T$. For example, $T$ is in hundreds for a grayscale image, whereas, $\delta \sim 1$. 
Therefore, it follows from \eqref{eps} that $\varepsilon \approx w(0) \delta/(2 T)$. On the other hand, from \eqref{LambW} and \eqref{series}, we have
\begin{equation*}
N_0 \approx \frac{q}{t - t^2} = \frac{e\lambda}{1-(q/e\lambda)},
\end{equation*}
where $t = q/e\lambda$ and $q=-\lambda + \log(1/\varepsilon)$. Since $|q| < e\lambda$,  
\begin{equation*}
\frac{1}{1-(q/e\lambda)} \approx 1+(q/e\lambda).
\end{equation*}
Therefore, $N_0 \approx e\lambda + q = (e-1)\lambda + \log(1/\varepsilon)$.

\section{Acknowledgements}

The authors thank Dr. Alessandro Foi and the anonymous reviewers for their useful comments and suggestions. 



\begin{thebibliography}{9}

 \bibitem{Chaudhury2015}  K. N. Chaudhury, ``Fast and accurate bilateral filtering using Gauss-polynomial decomposition,'' \textit{Proc. IEEE International Conference on Image Processing}, pp. 2005 - 2009, 2015.
 
\bibitem{Perona1990} P. Perona and J. Malik, ``Scale-space and edge detection using anisotropic diffusion,'' \textit{IEEE Transactions on Pattern Analysis and Machine Intelligence}, vol. 12, no. 7, pp. 629-639, 1990.

\bibitem{Tomasi1998} C. Tomasi and R. Manduchi, ``Bilateral filtering for gray and color images,'' \textit{Proc. IEEE International Conference on Computer Vision}, pp. 839-846, 1998.

\bibitem{bilat_application_book} S. Paris, P. Kornprobst, J. Tumblin, and F. Durand, \textit{Bilateral Filtering: Theory and Applications}, Now Publishers Inc., 2009.

\bibitem{Knaus2014} C. Knaus and M. Zwicker, ``Progressive image denoising,'' \textit{IEEE Transactions on Image Processing}, vol. 23, no.7, pp. 3114-3125, 2014.

\bibitem{CR2015} K. N. Chaudhury and K. Rithwik, ``Image denoising using optimally weighted bilateral filters: A SURE and fast approach,'' \textit{Proc. IEEE International Conference on Image Processing}, pp. 108-112, 2015.

\bibitem{Durand2002} F. Durand and J. Dorsey. ``Fast bilateral filtering for the display of high-dynamic-range images,'' \textit{ACM Transactions on Graphics}, vol. 21, no. 3, pp. 257-266, 2002.

\bibitem{Yang2009} Q. Yang, K. H. Tan, and N. Ahuja, ``Real-time $O (1)$ bilateral filtering,'' \textit{Proc. IEEE Conference on  Computer Vision and Pattern Recognition}, pp. 557-564, 2009. 

\bibitem{Paris2006} S. Paris and F. Durand, ``A fast approximation of the bilateral filter using a signal processing approach,'' \textit{Proc. European Conference on Computer Vision}, pp. 568-580, 2006.

\bibitem{Kamata2015} K. Sugimoto and S. I. Kamata, ``Compressive bilateral filtering," \textit{IEEE Transactions on Image Processing}, vol. 24, no. 11, pp. 3357-3369, 2015.
 
\bibitem{Porikli2008} F. Porikli, ``Constant time $O (1)$ bilateral filtering,'' \textit{Proc. IEEE Conference on Computer Vision and Pattern Recognition}, pp. 1-8, 2008.

\bibitem{Chaudhury2011} K. N. Chaudhury, D. Sage, and M. Unser, ``Fast $O(1)$ bilateral filtering using trigonometric range kernels,'' \textit{IEEE Transactions on Image Processing}, vol. 20, no. 12, pp. 3376-3382, 2011.

 \bibitem{Deriche1993} R. Deriche, ``Recursively implementing the Gaussian and its derivatives, \textit{Research Report}, INRIA-00074778, 1993.

\bibitem{Chaudhury2013}  K. N. Chaudhury, ``Acceleration of the shiftable algorithm for bilateral filtering and nonlocal  means," \textit{IEEE Transactions on Image Processing}, vol. 22, no. 4, pp. 1291-1300, 2013.

 \bibitem{Muller2006} J. M. Muller, \textit{Elementary Functions: Algorithms and Implementation}, Birkhauser Boston, 2006.
 
 \bibitem{errbilat}  S. An, F. Boussaid, M. Bennamoun, and F. Sohel, ``Quantitative error analysis of bilateral filtering," \textit{IEEE Signal Processing Letters}, vol. 22, no. 2, pp. 202-206, 2015.

 
 \bibitem{YDG2003}  C. Yang, R. Duraiswami, and N. A. Gumerov, ``Improved fast gauss transform,'' \textit{Technical Report} CS-TR-4495, UMIACS, Univ. of Maryland, College Park, 2003.

 \bibitem{prob} M. Mitzenmacher and E. Upfal. \textit{Probability and Computing: Randomized Algorithms and Probabilistic Analysis}, Cambridge University Press, 2005.
 
\bibitem{lamb} R. M. Corless, G. H. Gonnet, D. E. Hare, D. J. Jeffrey, and D. E. Knuth, ``On the Lambert W function," \textit{Advances in Computational Mathematics}, vol. 5, no. 1,  pp. 329-359, 1996.

\bibitem{Mcode} K. N. Chaudhury and S. Dabhade, \textit{Fast and Accurate Bilateral Filtering} (www.mathworks.com/matlabcentral/fileexchange/56158), MATLAB Central File Exchange, retrieved March 25, 2016.

\bibitem{dataset} http://www.imageprocessingplace.com/root\_files\_V3/image\_databases.htm.

\bibitem{Weiss2006} B. Weiss, ``Fast median and bilateral filtering,'' \textit{Proc. ACM Siggraph}, vol. 25, pp. 519-526, 2006.

\end{thebibliography}
\end{document}